\documentclass[conference]{IEEEtran}

\usepackage[square,numbers,sort&compress]{natbib}
\usepackage{multicol}
\usepackage{multirow}
\usepackage{xspace}
\usepackage{colortbl}
\usepackage[font=small, labelfont=bf]{caption}
\usepackage[table, dvipsnames]{xcolor}
\usepackage{amsmath}
\usepackage{amssymb}
\usepackage{booktabs}
\usepackage{lipsum}
\usepackage[export]{adjustbox}
\usepackage{float}
\usepackage{caption}
\usepackage[linesnumbered, ruled, vlined]{algorithm2e}
\usepackage{tikz}
\usepackage{scalerel}
\usepackage{url}
\usepackage{bbold}
\usepackage{tikz}
\usepackage{placeins}
\usepackage{subcaption}
\usepackage{tabularx}

\usepackage[colorlinks = true,
            linkcolor = magenta,
            urlcolor  = blue,
            citecolor = magenta,
            bookmarks = true, pagebackref]{hyperref}

\newcolumntype{C}[1]{>{\centering\arraybackslash}m{#1}}

\def\eqref#1{equation~\ref{#1}}

\def\1{\bm{1}}

\DeclareMathAlphabet{\mathsfit}{\encodingdefault}{\sfdefault}{m}{sl}
\SetMathAlphabet{\mathsfit}{bold}{\encodingdefault}{\sfdefault}{bx}{n}

\def\gA{{\mathcal{A}}}
\def\gB{{\mathcal{B}}}

\def\gD{{\mathcal{D}}}

\def\gM{{\mathcal{M}}}

\def\gS{{\mathcal{S}}}

\newcommand{\R}{\mathbb{R}}

\begin{document}
\thispagestyle{plain}
\pagestyle{plain}

\newcommand{\temp}[0]{{\color{red}\xspace TODO}\xspace}
\newcommand{\tempnote}[1]{{\color{red}\xspace [TODO #1]}\xspace}
\newcommand{\SysName}[0]{FastRLAP\xspace}
\newcommand{\tenth}{1/10$^\text{th}$\xspace} %
\newcommand{\MyPara}[1]{\vspace{1mm} \noindent\textbf{#1}}
\newcommand{\ours}[0]{\rowcolor{Cerulean!20}}
\newcommand\blfootnote[1]{%
  \begingroup
  \renewcommand\thefootnote{}\footnote{#1}%
  \addtocounter{footnote}{-1}%
  \endgroup
}

\newcommand*{\ditto}{---\texttt{"}---}
\newcommand{\colorme}[1]{\textcolor{#1}{#1}}
\newcommand{\human}[1]{\textcolor{gray}{#1}}
\newcommand{\practicecolor}[1]{\textcolor{NavyBlue}{#1}}
\newcommand{\pretraincolor}[1]{\textcolor{Mulberry}{#1}}
\newcommand{\onlinecolor}[1]{\textcolor{RedOrange}{#1}}
\newcommand{\env}[1]{\textbf{#1}}

\makeatletter%
\DeclareRobustCommand{\checkbold}[1]{%
 \def#1{0}%
 \edef\@tempa{\f@series}\edef\@tempb{\bfseries@rm}%
 \ifx\@tempa\@tempb%
  \def#1{1}%
 \fi%
 \edef\@tempb{\bfseries@sf}%
 \ifx\@tempa\@tempb%
  \def#1{1}%
 \fi}
\makeatother 
\newcommand\TalkingHead[1][]{\checkbold\tmp%
\scalerel*{\begin{tikzpicture}[line width={(1+0.67*\tmp)*pi*1mm},#1]
\path[use as bounding box] (-3.7,-4.2) rectangle (6.4,4.72);
\draw[xscale=-1] (-1,-4) to[out=50,in=-90,looseness=1.4] (-1,-3) to[out=-160,in=0]
   (-2,-3.2) to[out=180,in=-120] (-2.7,-2) to[out=120,in=-120] (-2.9,-1.5)
   to[out=120,in=-120] (-2.95,-1.1) to[out=60,in=-30] (-3.2,-0.9)
   to[out=150,in=-120] (-3,0.4) to[out=60,in=-90] (-3,1.2) 
   to[out=90,in=90,looseness=1.8] (3.3,1.2)
   to[out=-90,in=90,looseness=0.8] (2,-2)
   to[out=-90,in=150,looseness=0.8] (3.2,-4) -- cycle;
  \draw[shift={(pi,-pi/2)}] 
    foreach \X in {1,2,3} {(45:\X) arc[start angle=45,end
  angle=-45,radius=\X]};
\end{tikzpicture}}{B}}

\title{\huge \textbf{\SysName}: A System for Learning High-Speed Driving via \\ Deep RL and Autonomous Practicing}

\author{Kyle Stachowicz$^\dagger$, Arjun Bhorkar$^\dagger$, Dhruv Shah$^\dagger$, Ilya Kostrikov, Sergey Levine \\ UC Berkeley}

\makeatletter
\let\@oldmaketitle\@maketitle%
\renewcommand{\@maketitle}{\@oldmaketitle%
    \centering
    \vspace*{0.2em}
    \includegraphics[width=\linewidth]{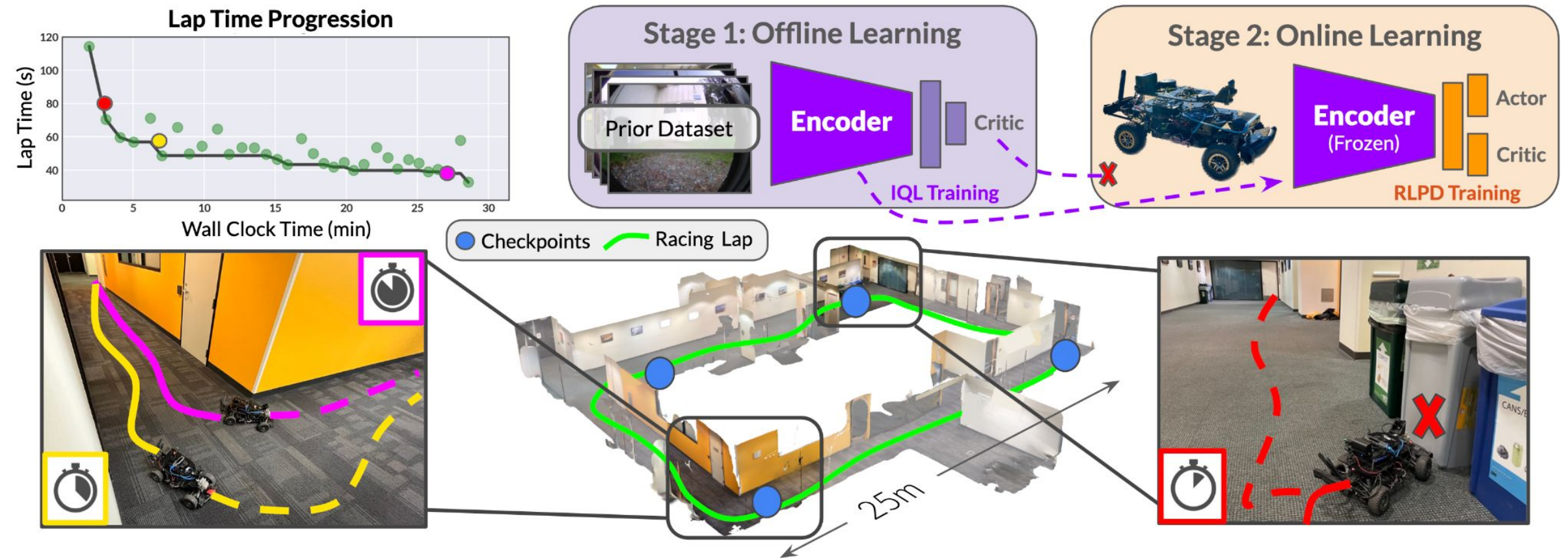}
    \captionof{figure}{\textbf{Fast reinforcement learning via autonomous practicing.} By pre-training the RL policy on diverse data (Stage 1), and deploying our autonomous practicing framework for continuous online improvements (Stage 2) in large real-world environments, the robot can autonomously navigate between sparse checkpoints (\textcolor{NavyBlue}{blue}), recovering from collisions during practice (\colorme{red}) and improve its driving behavior to maximize speed (\textcolor{Dandelion}{yellow} $\rightarrow$ \colorme{magenta}). \SysName can learn aggressive driving comparable to a human expert within 20 minutes of autonomous practice.} %
    \vspace*{-0.8em}
    \label{fig:teaser}
}
\makeatother

\maketitle
\IEEEpeerreviewmaketitle
\setcounter{figure}{1}

\begin{abstract}
We present a system that enables an autonomous small-scale RC car to drive aggressively from visual observations using reinforcement learning (RL). Our system, \SysName (\emph{faster lap}), trains autonomously in the real world, without human interventions, and without requiring any simulation or expert demonstrations. Our system integrates a number of important components to make this possible: we initialize the representations for the RL policy and value function from a large prior dataset of \emph{other} robots navigating in other environments (at low speed), which provides a navigation-relevant representation. From here, a sample-efficient online RL method uses a single low-speed user-provided demonstration to determine the desired driving course, extracts a set of navigational checkpoints, and autonomously practices driving through these checkpoints, resetting automatically on collision or failure. Perhaps surprisingly, we find that with appropriate initialization and choice of algorithm, our system can learn to drive over a variety of racing courses with less than 20 minutes of online training. The resulting policies exhibit emergent aggressive driving skills, such as timing braking and acceleration around turns and avoiding areas which impede the robot's motion, approaching the performance of a human driver using a similar first-person interface over the course of training.
\blfootnote{$^\dagger$ Equal Contribution. Code and videos at  \href{https://sites.google.com/view/fastrlap}{\bf sites.google.com/view/fastrlap}. }
\end{abstract}

\section{Introduction}
\label{sec:intro}

High-speed vision-based navigation presents a range of challenges: aside from the usual difficulties associated with collision-free navigation, it requires controllers that can account for both the vehicle's dynamics and the perceived obstacles (Fig.~\ref{fig:challenges}). Learning-based methods offer a particularly appealing way to approach such challenges, as they can directly learn the relationship between perception and vehicle dynamics and in principle capture high-performance driving behaviors. One way that prior work has approached such domains is via imitation learning, acquiring end-to-end skills from expert demonstrations~\citep{bojarski2016end2end, bansal2018chauffeurnet}. However, if our aim is to maximize performance, we might instead prefer to directly adapt the navigational strategy to the vehicle \emph{autonomously}. By learning from \emph{autonomous} experience in an environment, reinforcement learning (RL) can enable this, analogously to progress in other domains such as games and robotic manipulation, where policies trained via RL have even exceeded human-level performance~\citep{mnih2013playing, schrittwieser2020mastering, gu2017deeprl, kroemer2021survey}.

But the autonomous setting presents major challenges for RL: unlike other domains, it is impossible to reset the system to a random state, and so the learning process is highly dependent on the system's ability to make continual progress without getting \emph{stuck}. Starting the training process from a randomly-initialized policy and learning solely via trial-and-error would likely result in catastrophic failure. Instead, the RL-based system should train automatically, without supervision, not just improving its policy performance but also smoothly recovering from failures or collisions. The goal of this paper is to address these challenges and understand how RL can be applied to autonomously learn high-speed driving from vision.

\begin{figure*}[ht!]
    \centering
    \includegraphics[width=\linewidth]{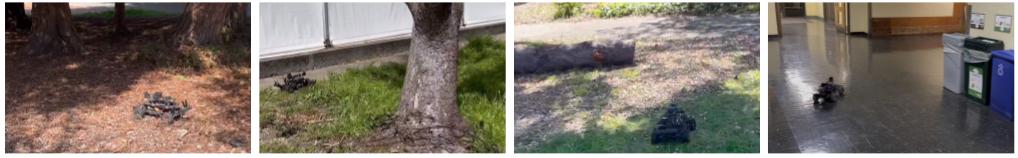}
    \caption{FastRLAP learns fast driving policies for a \tenth-scale vehicle operating in diverse indoor and outdoor environments with challenging terrains in tens of minutes, using a combination of offline pre-training and online reinforcement learning.}
    \label{fig:teaser_envs}
\end{figure*}

Directly applying RL methods in the real world with high-dimensional observations is very difficult: online RL algorithms require large numbers of samples, and trial-and-error learning can drive to robot into unsafe or irrecoverable situations, necessitating frequent human interventions. As an alternative, we could learn entirely from offline data~\citep{lange2012batch, levine2020offline}, but this might yield suboptimal policies, as the desired behavior (e.g., driving aggressively in tight spaces) might not be well-supported under the prior data. Therefore, our approach combines online RL training with a set of pre-training and initialization steps that aim to maximally transfer prior knowledge to \emph{bootstrap} real-world RL. Specifically, we aim to use prior data to learn a useful representation of visual observations that captures driving-related features, such as free space and obstacles, while also adapting online to the target domain.

To this end, we design a high-performance system for \textbf{Fast} \textbf{R}einforcement \textbf{L}earning via \textbf{A}utonomous \textbf{P}racticing. \SysName (\TalkingHead~\emph{faster-lap}), which uses an automatic goal curriculum to guide a goal-conditioned RL policy to quickly adapt to the target environment and improve its performance over multiple laps, without requiring human interventions (see Fig.~\ref{fig:teaser}).
We then leverage data-efficient online RL, bootstrapped with just a single low-speed demonstration of the track, to rapidly learn a policy that can drive aggressively in challenging environments (see Fig.~\ref{fig:teaser_envs}). This stage takes under 20 minutes, accelerated by pre-trained representations and enabled by a sample-efficient online RL procedure that effectively uses them.

The primary contribution of this work is \SysName, a system for autonomous learning of vision-based navigation that leverages diverse prior data and improves by practicing autonomously.
We demonstrate our \SysName in challenging environments on a custom \tenth-scale RC car modified for real-world online RL. \SysName can autonomously practice and learn aggressive maneuvers using a novel autonomous practicing framework, improving by up to 40\% over the demonstration lap and achieving performance close to a human expert.
Notably, the online training phase typically takes less than 20 minutes (as little as 5 minutes!), depending on the size of the environment. During this time, the robot learns aggressive maneuvers, drifting, and maintaining a racing line, without any expert demonstrations. The training requires no human interventions and is fully autonomous.
To the best of our knowledge, \SysName is the first instantiation of a vision-based mobile robotic system that uses model-free RL to autonomously practice high-speed driving maneuvers and improve online in the real world. %

\section{Related Work}
\label{sec:relatedwork}

Leveraging prior data to bootstrap online learning has been widely studied in the context of supervised learning~\cite{ross2011reduction}, representation learning~\cite{yarats2021improving}, continual learning~\cite{thrun1995lifelong, bengio2009curriculum, florensa2018automatic, Matiisen2020teacher, sukhbaatar2018intrinsic}, and RL~\cite{levine2020offline}. Offline RL has proven particularly powerful due to its ability to learn actionable representations directly from existing large datasets,
and some works have studied how it can be combined with fine-tuning through online interaction~\cite{nair2020awac, villaflor2021finetuning, Xie2021PolicyFB, lee2021offlinetoonline}. This has enabled a variety of robotic systems that can leverage a combination of offline data and online interaction to perform real-world tasks~\cite{walke2022dont, kumar2022pre, gurtler2023benchmarking}. However, most such experiments focus on robotic manipulation or other scenarios that can evaluated in controlled workspaces. Instead, \SysName aims to perform fully autonomous practicing in unstructured environments, spanning over 120 meters. To address such real-world domains, \SysName incorporates pre-training data from a variety of environments and different robots, and employs a combination of diverse but less relevant demonstrations to enable effective bootstrapping. Additionally, \SysName integrates RL with a high-level practicing pipeline to enable greater autonomy during training.

Learning high-speed navigation has been approached in various ways. Typically, these systems rely either on highly accurate position information to define states~\cite{funke2012audi, Keivan2013RealtimeSC, Williams2016AggressiveDW, rosolia2020learning}, localize visual observations relative to a high-fidelity \emph{global map}~\cite{drews2017aggressive, Drews2019vision}, or operate via behavior cloning against some privileged expert which itself can access ground-truth state and mapping~\cite{pan2020imitation}. This can be prohibitive in unstructured environments, where (i) onboard state estimates can be highly inaccurate due to poor localization via noisy odometry or GPS measurements, and (ii) generating a high-fidelity map can be difficult or impossible. In contrast, \SysName learns aggressive driving behavior directly from vision, using only a coarse sequence of checkpoints, and can improve its behavior by self-practice without using privileged state information.

\begin{figure}
    \centering
    \includegraphics[width=\columnwidth]{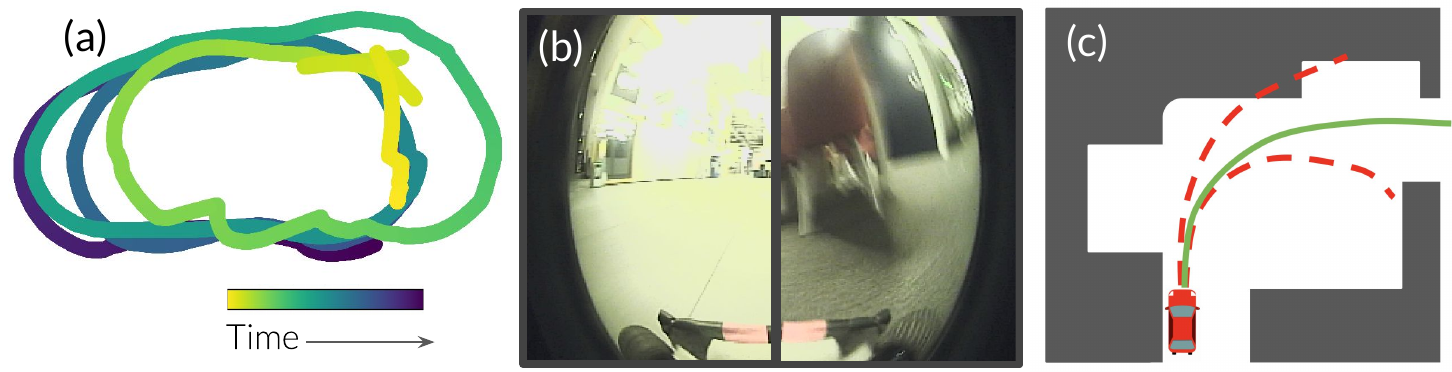}
    \caption{High-speed visual navigation faces challenges due to sensing, perception, and dynamics: (a) noisy odometry and localization errors across multiple laps, (b) overexposure and motion blur in the visual observations, and (c) over-/under-steer due to complex dynamics.}
    \label{fig:challenges}
\end{figure}

Prior successes in learning visual navigation for ground and aerial robots often involves either learning from large-scale simulated data~\cite{loquercio2020deep, loquercio2021learning, fuchs2021granturismo, Gervet2022NavigatingTO}, passive data~\cite{chang2020youtube}, human interventions~\cite{kendall2018learning, kahn2021land} or leveraging real-world data from other robots~\cite{shah2022gnm}. However, simulating off-road environments can be extremely challenging due to complex relationships between perception, vehicular dynamics, and terrain (Fig.~\ref{fig:challenges}), human interventions are time-consuming and expensive, and aggressive driving maneuvers tend to be closely adapted to the specific environments, calling for learning directly from \emph{on-task} data. The closest prior work~\cite{shah2022offline} uses offline RL for off-road driving, but does not support mechanisms to practice autonomously or adapt its behavior on-the-fly. We present the first navigation system that combines offline pre-training with fast, online RL training that can improve with experience and adapt to individual real-world environments autonomously.

A number of prior papers have studied the problem of autonomous real-world RL, often through the lens of safety or reset-free training, where the need for human interventions during training is minimized~\cite{Han2015learning, Richter2017SafeVN, eysenbach2017leave, zhu2020ingredients, lu2021resetfree, sharma2022autonomous}, and has been applied to robotic manipulation~\cite{gupta2021mtrf, walke2022dont}, quadruped locomotion~\cite{Ha2020LearningTW}, and mobile manipulation~\cite{sun2022relmm}.
We draw inspiration from these ideas to build an aggressive navigation system that uses a finite state machine to \emph{practice} driving around a circuit. \SysName scales to novel environments, driving around 100+ meter courses, and can continually improve its performance.

\section{Problem Formulation}
\label{sec:prelims}

The objective of our high-speed visual navigation task is to drive through a race course, defined as a sequence of position \emph{checkpoints} $\{c_i\}$, in the minimum possible time.
We frame this task as a Markov decision process (MDP) $\gM (\gS, \gA, p, r)$, where $\gS$ denotes the state space comprising of states  $s = (V, v, \omega, \alpha, g, a_{\text{prev}})$.
Here, $V \in \R^{128\times 128 \times 3 \times 3}$ is a stacked sequence of the last 3 RGB images; $v, \omega, \alpha \in \R^3$ denote the linear velocity, angular velocity, and linear acceleration; the goal $g$ is provided as a relative vector to the next checkpoint, written as a 2D unit vector and a distance; $a_{\text{prev}}$ is the previous action. All measurements are specified relative to the robot's internal reference frame: the policy does not require information in any fixed external frame of reference. The action space $\gA$ is specified by motor velocity targets corresponding to throttle and steering actions. Note that we do not allow the throttle command to be negative, the policy can only drive the robot forwards. $p$ denotes the unknown transition dynamics, and $r$ is a reward function corresponding to reaching $c_i$ quickly (described in Sec.~\ref{sec:methods-online}).

To make training in the real world practical and automated, we formulate this problem in the context of autonomous RL~\cite{sharma2022autonomous}, where the robot is not provided with periodic resets or interventions when it collides with an obstacle or gets stuck, and needs to automatically recover on its own. The robot must \emph{practice} driving around the lap to fully master the task and improve its performance over time \emph{without} any human interventions. We measure the robot's performance along three metrics: (i) time to complete the first lap, (ii) final and median lap times over the course of training, and (iii) the mean number of collisions per lap over the course of training.

\section{Autonomous Practicing with RL}
\label{sec:methods-main}

Our system for learning high-speed driving, \SysName, aims to enable autonomous practicing and sample-efficient end-to-end RL in the real world. \SysName has three components: a simple high-level finite state machine (FSM) for autonomous practicing, a pre-trained representation of visual observations, and a sample-efficient RL algorithm for online learning (see Fig.~\ref{fig:teaser}). The FSM (shown in \practicecolor{blue}) serves the dual purpose of selecting the next checkpoint for the online RL policy and automatically recovering from collisions, enabling autonomous practicing in the real world. The online RL policy (shown in \onlinecolor{orange}) is trained \emph{online} in the real world to reach goals commanded by the FSM, and continually improves to learn aggressive driving maneuvers. To provide for compute- and sample-efficient training of the online RL policy, we bootstrap it with an offline representation of navigation-specific visual features trained from prior data (shown in \pretraincolor{purple}).

\subsection{Autonomous Practicing and Goal Checkpoint Selection}
\label{sec:methods-fsm}

In the autonomous learning setting, the robot is expected to learn in the environment without any episodic ``resets'' or human interventions. In early stages of training, the RL policy may make mistakes that lead to irrecoverable states, such as collisions. Without the help of a reset, the robot would get stuck forever and the learning algorithm may become degenerate due to collapse in the state distribution~\cite{zhu2020ingredients}. To overcome this, we use a simple FSM that switches between collision recovery and commanding a relevant goal checkpoint to the RL agent.

This FSM serves a dual purpose. When the RL policy reaches a goal checkpoint, as measured by a low-fidelity localization estimate (e.g., from visual-inertial odometry or GPS),
the FSM commands a new goal corresponding to the next checkpoint in the course sequence $\{c_i\}$. This forces the learner to practice reaching all of the checkpoint goals on the race course sequentially. If the RL policy lands itself into an irrecoverable state (e.g., experiences a collision or becomes stuck, see Sec.~\ref{sec:system} for implementation details),
the FSM commands an automatic recovery policy to rescue the robot and provides a ``pseudo-reset''. We use a very simple scripted recovery policy to perturb the robot's state, which selects a random steering angle and drives backward for a short distance. Although other approaches such as using exploratory policies~\cite{zhu2020ingredients} or learning a recovery policy~\cite{eysenbach2017leave} have also been studied, we found this very simple strategy to be sufficient to allow the online RL procedure to learn directly in the real world without human interventions.

\begin{algorithm}[h]
\SetKwBlock{Robot}{On Robot}{}
\SetKwBlock{Desktop}{On Workstation}{}
\KwData{Prior navigation dataset $\gD$, slow demo lap $\gB_{\text{slow}}$}
\textbf{Keys:} \pretraincolor{Pre-Training}, \practicecolor{Practicing}, \onlinecolor{Online RL} \\
\caption{\SysName for Autonomous Practicing}
\label{alg:main}
\While{Encoder is not converged}{
\pretraincolor{
    $s, a, s', \textrm{idx} \gets$ LoadData($\gD$) \\
    $g \gets$ LoadFutureData($\gD$, idx + RandomOffset()) \\
    $r \gets $ ComputeReward($s$, $a$, $g$) \\
    Train$_\text{IQL}$($(s, g), a, r, (s', g)$) \label{algline:iql-update}   \\
}}
\While{True}{
    \Robot{
    \label{algline:robot}
        $s \gets $Observe()\\
        \practicecolor{\If{$s$ \textnormal{near} $g$}{
            $g \gets \text{NextCheckpoint}(g)$\\
            \label{algline:goal-setting}
        }}
        \onlinecolor{
        $r \gets $ ComputeReward($s_{\textrm{prev}}$, $a_{\textrm{prev}}$, $g$)\\
        SendToWorkstation($s_{\textrm{prev}}$, $a_{\textrm{prev}}$, $r$, $s$, $g$)\\
        \label{algline:send-to-trainer}
        $a \sim \pi(\phi(s_{\textrm{image}}), s_{\textrm{proprio}}, g)$\\ \label{algline:online-policy} }
        Actuate($a$)\\
        \practicecolor{\If{Collision \textnormal{or} Stuck}{
            Execute recovery policy
            \label{algline:collision-recovery}
        }}
    }
    \Desktop{
    \label{algline:desktop}
        \onlinecolor{ReceiveFromRobot($\gB$)\\
        $b$ $\gets$ Sample($\gB$), $b_{\text{slow}} \gets$ Sample($\gB_{\text{slow}}$) \label{algline:slow-sample} \\
        $\pi, Q \gets $Train$_\text{RLPD}$($\pi, Q, b, b_{\text{slow}}$)\\
        \label{algline:rlpd-update}     
    }}
}
\end{algorithm}

\subsection{Online RL Training}
\label{sec:methods-online}

The objective of the low-level policy $\pi$ is to reach the goal checkpoints commanded by the FSM in the minimum possible time without colliding or getting stuck. Given the visual observations and a goal vector $\vec g$ from the FSM, $\pi$ must parse the high-dimensional observations to understand its
contents and plan a high-speed trajectory through it without colliding with obstacles or getting stuck. It is important to note that the goal checkpoints $c_i$ are typically beyond line-of-sight (e.g., Fig.~\ref{fig:teaser}, \practicecolor{blue}), up to 40 meters away,
and navigating between them requires the robot to learn a representation of the environment layout, correlating visual observations with possible \emph{racing lines} that the robot could take to maintain high speed.

We design a simple reward function $r$ that prioritizes maintaining maximum instantaneous velocity towards the next goal checkpoint, while also avoiding irrecoverable states that require the FSM to trigger the recovery policy, slowing down the robot's total lap time. We primarily define the reward as \emph{speed-made-good}: the component of the instantaneous velocity in the direction facing the next checkpoint. We additionally add a penalty for becoming stuck (defined as failing to move despite commanding non-zero throttle) and colliding:
\begin{equation}
    r(s, a) = \vec v \cdot \frac{\vec g}{\lVert \vec g \rVert} - C_{\textrm{stuck}}\mathbb{1}_{\textrm{stuck}} - C_{\textrm{collide}}\mathbb{1}_{{\lVert a \rVert} > A}\lVert a \rVert.
    \label{eq:reward}
\end{equation}
Here, $\vec v$ and $\vec g$ are the observed velocity and relative goal coordinates (included in the state observation $s$), $a$ is lateral acceleration, $\mathbb{1}$ is the indicator function to detect irrecoverable states, and $C_\text{stuck},C_\text{collide},A > 0$ are constants.

To maximize the above reward and continually improve the robot's lap times, the system must learn from interactions with the environment in practice laps, using a \emph{batch} of new interactions to update its learned behavior using off-policy RL~\cite{mnih2013playing, fujimoto2019off}. Such approaches benefit greatly by performing multiple training steps for each environment step, known as the update-to-data (UTD) ratio: a large UTD leads to efficient learning, but often suffers from overfitting. To overcome this, we train our policies with RLPD~\cite{anon2023RLPD}, a data-efficient off-policy RL algorithm that trains an ensemble of critics to avoid catastrophic overestimation~\cite{chen_randomized_2021} and can learn quickly using a combination of online interactions and a small amount of suboptimal, \emph{on-task} data.

We obtain this \emph{on-task} data by collecting a single \emph{slow} lap in the target environment, similar to how a racing driver might perform a reconnaissance lap at low speed to familiarize themselves with the course before attempting laps at high speeds. While this data is very limited (under a minute in most environments) and does not contain fast-driving behaviors, prior work has shown that it can significantly accelerate online learning by avoiding critic collapse in early stages of training~\cite{anon2023RLPD}. During the online training, we sample 50\% of each training batch from this low-speed data, interleaved with 50\% of data collected online. We found this to be critical to the efficiency of our system in our evaluations (Sec.~\ref{sec:results-sim}).

However, our system aims to learn effective high-speed navigation skills in as little as 10-20 minutes. At such low training times, the process is constrained not only by the robot's ability to collect data, but also by the \emph{computational constraints} of training the neural network, so improvements in computationally efficiency actually translate directly into faster learning with higher UTD ratios. Therefore we combine the strengths of data-efficient online learning with a powerful pre-trained representation of visual observations to enable computation-efficient training.

\subsection{Representation Learning with Offline RL}
\label{sec:methods-offline}

When training image-to-action policies, end-to-end RL allows gradients from the control objective to optimize the encoder. This results in a task-specific encoder that produces features that are most relevant to the agent's task, rather than general features (e.g., features necessary for classification or video prediction). Unfortunately, training directly on full images is very computationally expensive and unacceptably reduces the UTD ratio. Ideally, we would prefer to pre-train some encoder to produce task-relevant features \emph{without} requiring new environment interactions, and then freeze the encoder during online training.

We address this by training the encoder with offline RL on an existing large-scale dataset with a similar (but not identical) objective.
In particular, we use RECON~\cite{shah2021rapid}, a large-scale navigation dataset collected by manually driving a Clearpath Jackal UGV outdoors at low-speeds. This dataset contains navigation trajectories from many environments and an entirely different robot, which importantly does \emph{not} include aggressive high-speed driving. Thus, the role of pre-training is not to teach the robot how to drive quickly and efficiently, but only to extract a navigation-relevant representation to simplify the online learning problem. The aggressive, high-speed driving behaviors necessary to solve the desired task must be learned through practice in the real world, building on this pre-trained foundation.

We apply goal-conditioned offline RL by selecting a 1:1 mixture of random goals and goals from the robot's future trajectory in this dataset, and use Implicit Q-Learning \cite{kostrikov2021offline} to train a critic network (illustrated in Fig.~\ref{fig:teaser}, \pretraincolor{purple}). We then take the learned encoder, which now encodes features relevant to the navigation task, and freeze it for training the aggressive driving policy and critic (\onlinecolor{orange}) as illustrated in Sec.~\ref{sec:methods-online}.

\subsection{Summary}
\label{sec:method-summary}

Algorithm~\ref{alg:main} summarizes the \SysName autonomous training framework. We use a diverse offline navigation dataset $\gD$ to \pretraincolor{pre-train} a navigation-relevant representation of visual observations using a goal-conditioned RL objective optimized with IQL (L\ref{algline:iql-update}). We freeze the encoder trained with this process and and use it to encode visual observations for the actor and critic for the online learning phase (see Fig.~\ref{fig:teaser}, \onlinecolor{orange}). The reward function for both the pre-training and online RL tasks is described by Eqn.~\ref{eq:reward}.

During deployment in a previously unseen environment, the practicing FSM (\practicecolor{blue}) serves the dual purpose of commanding the next goal checkpoint to the low-level policy $\pi$ (L\ref{algline:goal-setting}), and automatic collision recovery if the robot is stuck or in collision (L\ref{algline:collision-recovery}). To enable fast training, the inference is split between the robot and a remote workstation with low-latency communication between them (see Sec.~\ref{sec:system} for details).

The robot runs fast inference of the trained policy (L\ref{algline:online-policy}) at 10Hz, and sends batches of online experience to the workstation (L\ref{algline:send-to-trainer}) to asynchronously update the actor and critic networks using RLPD (L\ref{algline:rlpd-update}) as quickly as possible. This process enables \SysName to learn aggressive driving behavior from as little as 10-20 minutes of online experience.

\section{System Design for Online Learning}
\label{sec:system}
\begin{figure}
    \centering
    \includegraphics[width=0.9\columnwidth]{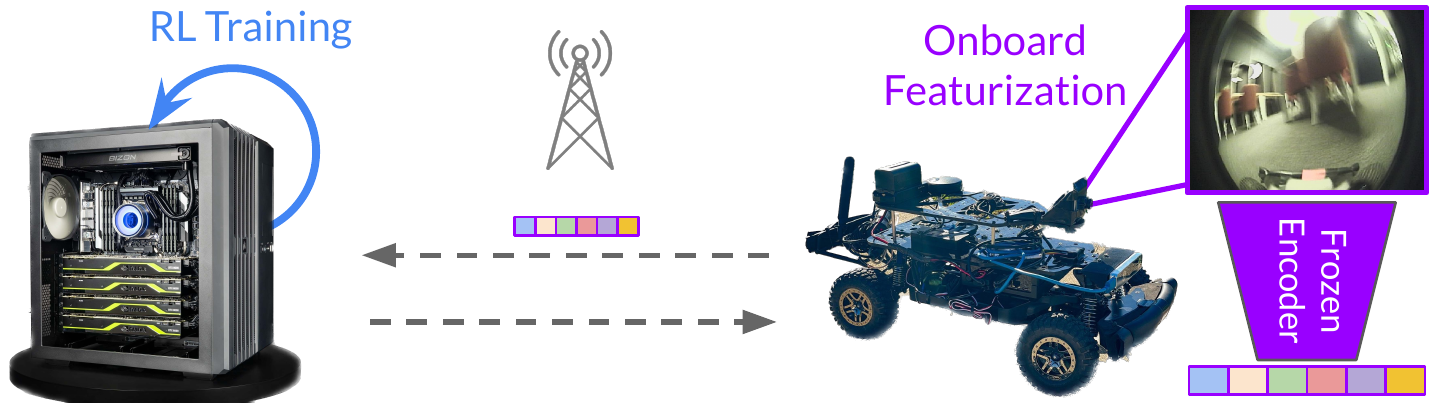}
    \caption{For fast training, we featurize the visual observations using a pre-trained encoder onboard the robot (right) and run RL training on these embeddings on a wirelessly connected workstation (left).}
    \label{fig:system-design}
\end{figure}

We instantiate our autonomous practicing system on a bespoke \tenth-scale autonomous rally car for high-speed navigation. While the base chassis was inspired by similar platforms~\cite{racecar, kelly2020tenth}, the task of autonomous practicing with online learning adds a number of challenges that necessitate hardware and networking modifications. We describe these differences on top of a commercially available Traxxas Slash 4$\times$4 Ultimate RC car (Fig.~\ref{fig:system-design}). %

\MyPara{Sensing:}
Since our high-speed system operates directly on visual observations, we use a forward-facing PCB camera with a fisheye lens to obtain a low-latency stream of 128$\times$128 RGB images with minimal motion blur. Since the learned policy requires coarse relative position estimates to intermediate checkpoints/subgoals, our system also requires a low-fidelity state estimator.

\textit{Indoor state estimation:}
Relying on wheel speeds and onboard IMU for local odometry is impractical due to wheel slippage. To estimate state in indoor environments, we use a RealSense T265 tracking camera to provide local visual-inertial odometry estimates for the positions of the robot and intermediate checkpoints. We orient the T265 to face upwards to minimize localization issues faced when colliding with an obstacle.

\textit{Outdoor state estimation:}
The VIO-based T265 system gives poor performance in outdoor environments. Instead, we use a state estimate from a ublox ZED-F9P GPS receiver mounted onboard the robot. Due to the difficulty of estimating ground-truth heading from only local sensor information, and the poor performance of magnetometers in close proximity to motors, we used an Extended Kalman Filter to estimate the heading $\theta$. We fuse local wheel odometry (forward velocity $v_w$) with absolute GPS velocity $\vec v$ and IMU angular velocity $\omega$ using the following dynamics and measurement models:
\[\theta_{t+1} = \theta_t + \omega \Delta t \hspace{30pt}\vec v = (v_w\cos \theta, v_w\sin\theta)\]

\MyPara{Compute:}
Following F1TENTH~\cite{kelly2020tenth}, we use an NVIDIA Jetson Xavier NX for onboard compute. We process visual observations onboard using a pre-trained encoder (Sec.~\ref{sec:methods-offline}), and offload the RL training to a desktop workstation over a WiFi network or LTE (see Fig.~\ref{fig:system-design}). This has the dual benefits of low communication latency (since we only communicate low-dimensional features, under 100kB/s) and fast training (since only a subset of the layers need to be updated). To achieve a high UTD ratio, we implement our training algorithm in JAX~\cite{jax2018github}. Using the just-in-time compiler to combine many update steps into a single optimized XLA function, we are able to increase our update rates to $\sim$800 actor updates per second. When combined with the on-robot featurization, this represents a $\sim$10x increase in throughput compared to training without these optimizations.

\MyPara{Actuation:}
The Markov assumption requires that the transition dynamics $p$ only depend on the current state and action. However, typical ``sensorless'' motors used in RC cars exhibit \emph{cogging}, a stuttering behavior that depends heavily on unobservable variables such as rotor positions. This occurs frequently when steering sharply or starting from a standstill, actions that the RL agent takes often during exploration. We upgrade the system to use a ``sensored'' motor to provide closed-loop startup sequencing without cogging. While the mechanical top speed of our system is nearly 30m/s, we cap the operational speed in indoor environments to 3.5m/s and in outdoor experiments to 4.5m/s for safety.

\begin{figure*}
    \centering
    \includegraphics[width=\textwidth]{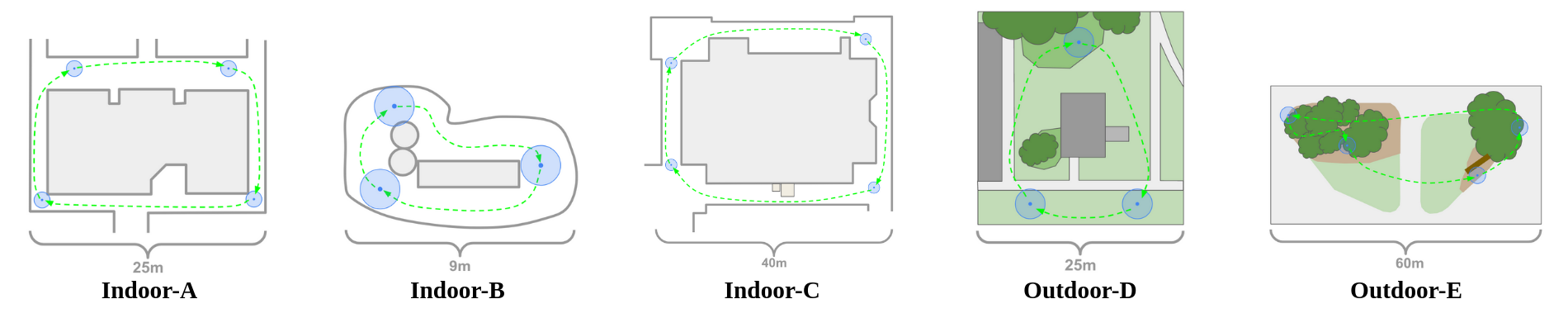}
    \caption{Schematics for each test environment in the real-world. FastRLAP is able to learn to drive real-world courses of varying size in both indoor and outdoor settings. \emph{Note that these schematics are not available to our system and are shown shown only for illustration.}}
\end{figure*}

\MyPara{Action continuity:}
The closed loop PID and servo actuators tend to act as a low-pass filter on their targets, smoothing out high-frequency changes. When sampling uncorrelated actions from an output distribution, this will result in a low signal-to-noise ratio between the commanded target actions and measured velocity observations, making critic learning via temporal difference learning difficult. %
To overcome this, we enforce \emph{continuity} in the policy's outputs by constraining them to be near the previous action by modifying the action space: (i) instead of the standard $\tanh$ activation to limit the action space in $[-1, 1]$ (used by actor), we use a \emph{shifted} $\tanh$ that limits it to the range $[a_\text{prev} - \delta, a_\text{prev} + \delta]$, i.e., near previous action, bounded by $\delta>0$, and (ii) we append the previous action to the observed state. Please see the supplemental material for further details.

\MyPara{Detecting blocked states:}
Our autonomous practicing system uses a recovery policy when the robot fails to make forward progress (e.g., collision or stuck). We detect collisions via a simple heuristic condition: at each timestep, we compare the lateral acceleration of the robot to a threshold value; if that threshold is violated, the robot is in collision and a penalty is applied according to Eqn.~\ref{eq:reward}. The ``stuck'' condition is detected using local odometry: if the robot has not moved at least 0.5m in the past three seconds, it is considered to be stuck and a pseudo-reset is performed.

\begin{figure}
    \centering
    \includegraphics[width=\columnwidth]{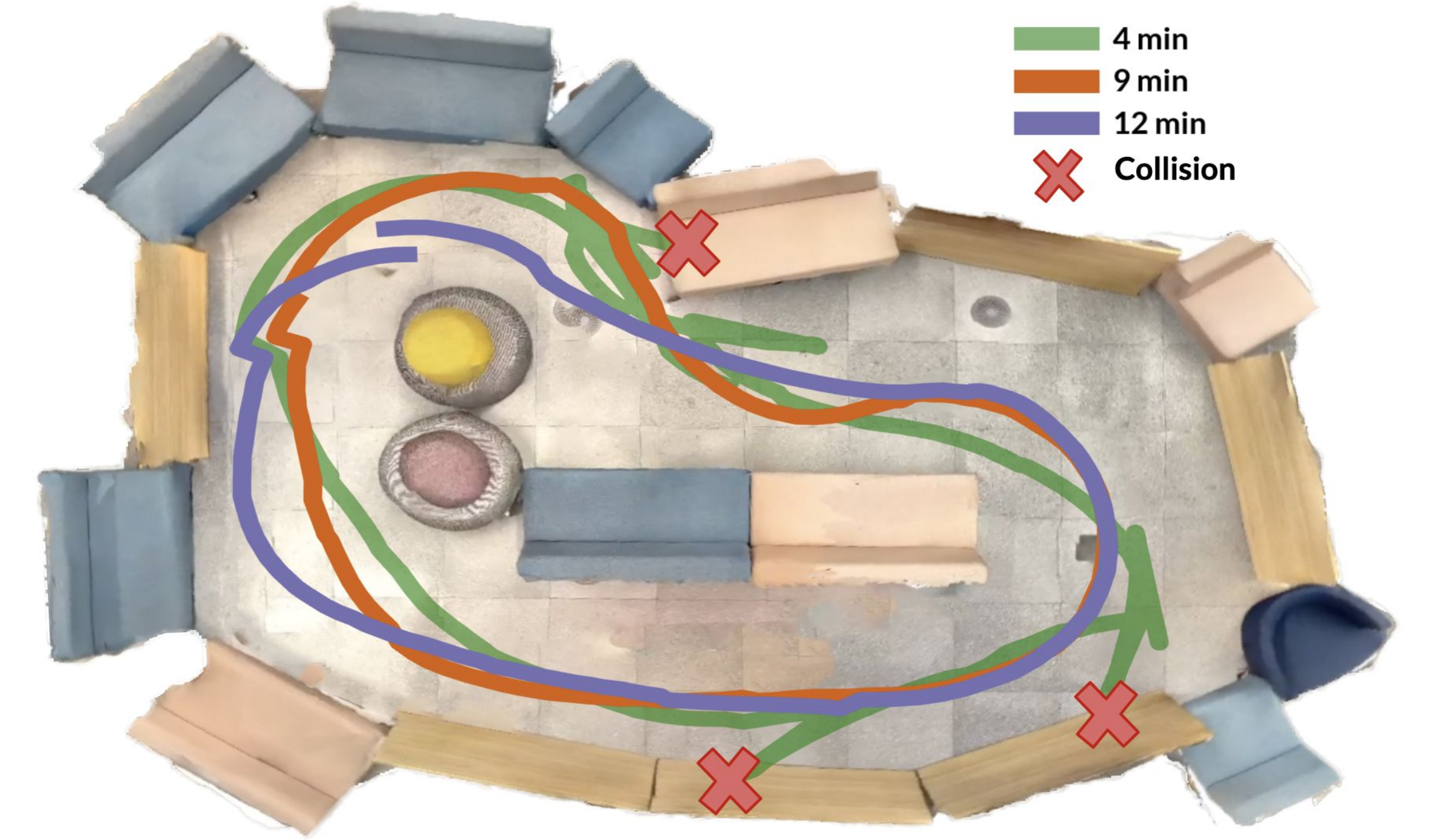}
    \caption{Sample trajectories of \SysName practicing in a real-world indoor environment. Starting from a pre-trained representation, \SysName often collides into obstacles (green) and learns from it's mistakes to learn collision-free navigation (orange). By directly minimizing lap times, \SysName discovers a smooth racing line (purple), matching human driving performance. \emph{Note that the 3D scan is not availably to our system and is shown only for illustration.}}
    \label{fig:real-evolving}
\end{figure}

\begin{figure*}

    \centering
    \begin{tabular}{C{0.2cm}C{3cm}C{4.7cm}|C{4.7cm}C{3cm}C{0.2cm}}
        \rotatebox{90}{\small \env{Indoor-A}} & \includegraphics[width=3cm, trim={0px 10px 0px 0px}, clip]{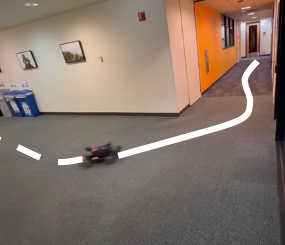} & \includegraphics[width=4.7cm]{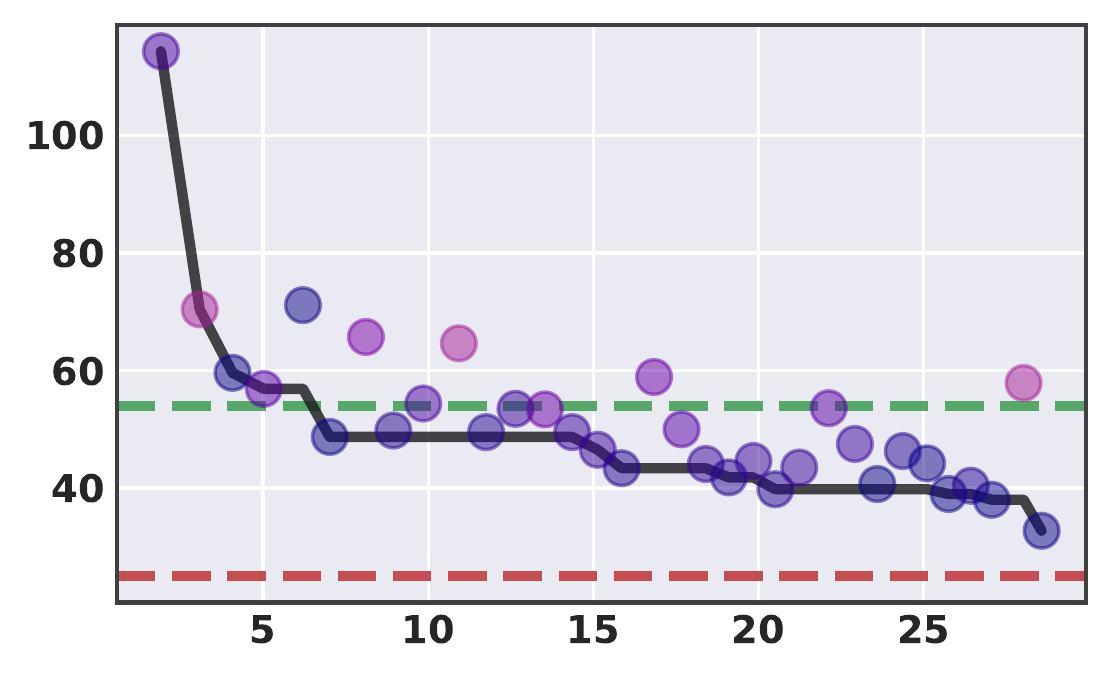}
        & \includegraphics[width=4.7cm]{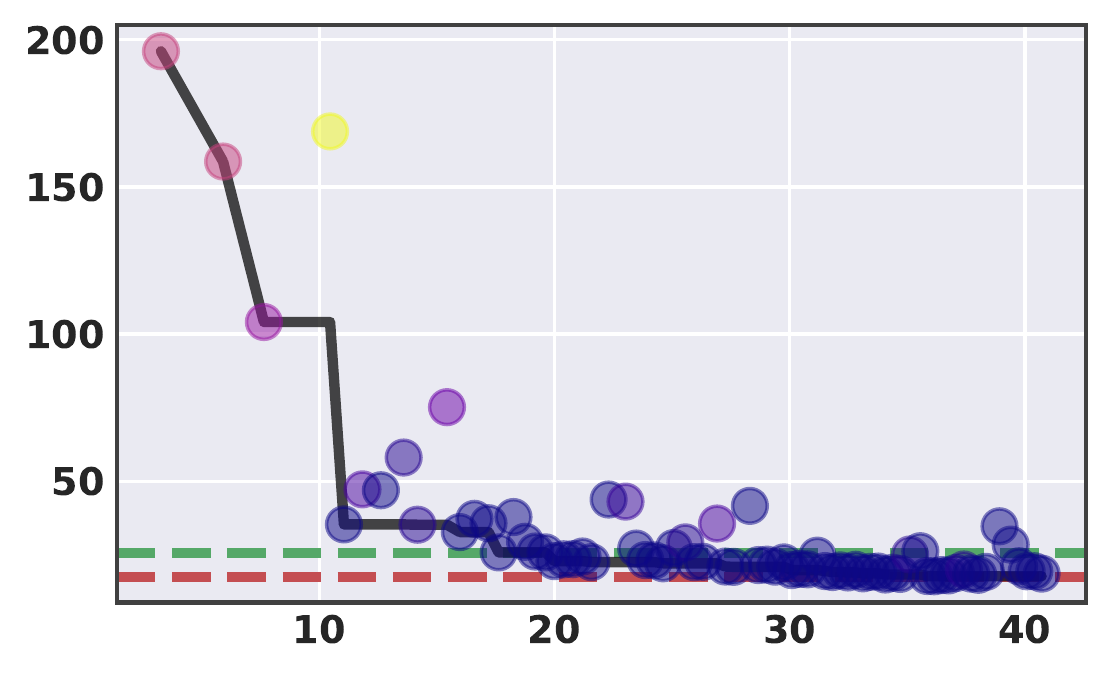} & \includegraphics[width=3cm, trim={20px 0px 40px 0px}, clip]{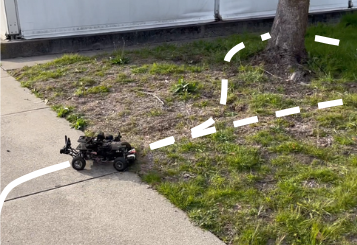} & \rotatebox{90}{\small \env{Outdoor-D}} \\
        
        \rotatebox{90}{\small \env{Indoor-B}} & \includegraphics[width=3cm, trim={0px 0px 0px 30px}, clip]{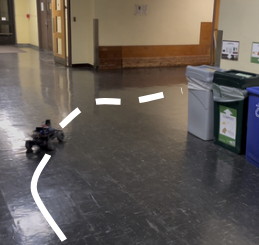} & \includegraphics[width=4.7cm]{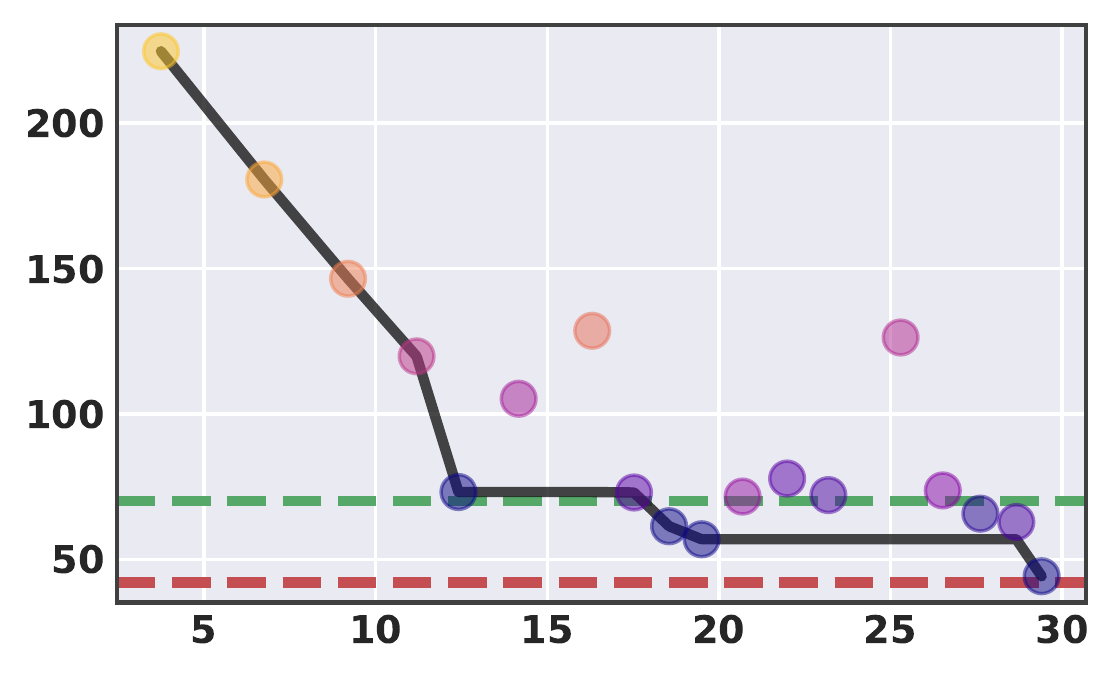}
        & \includegraphics[width=4.7cm]{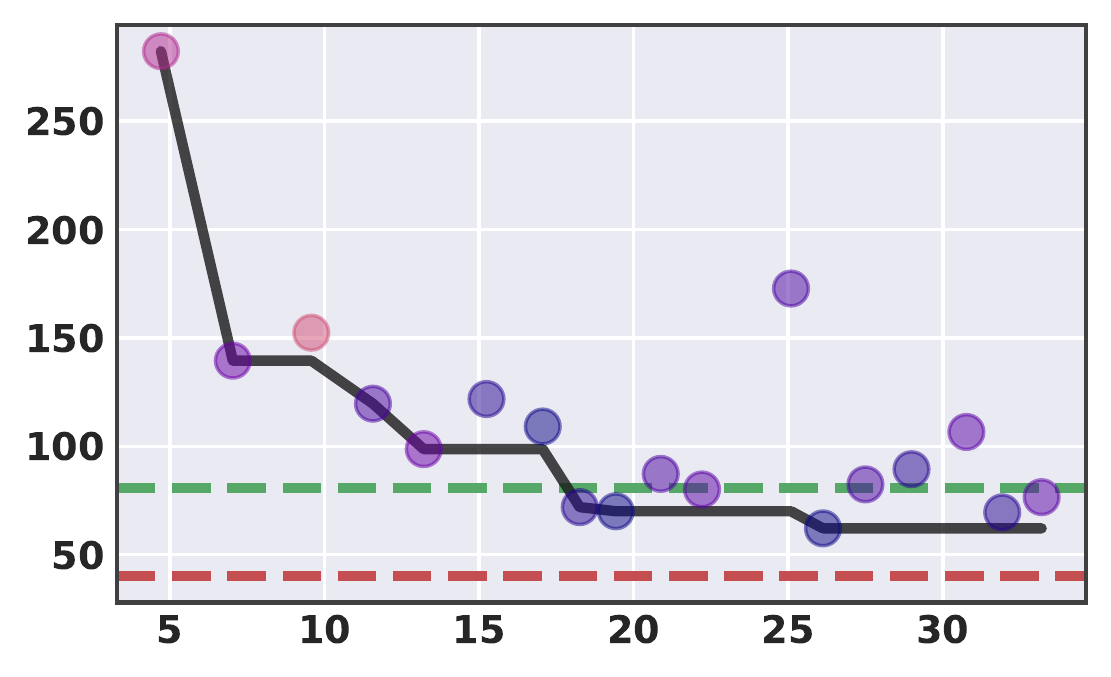} & \includegraphics[width=3cm, trim={0px 0px 50px 0px}, clip]{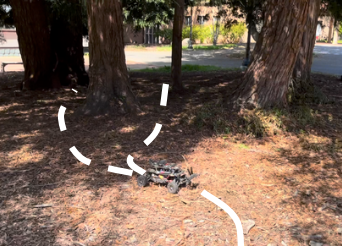} & \rotatebox{90}{\small \env{Outdoor-E}} \\
        
        \rotatebox{90}{\small \env{Indoor-C}} & \includegraphics[width=3cm, trim={0px 0px 0px 20px}, clip]{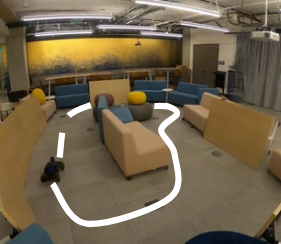} & \includegraphics[width=4.7cm]{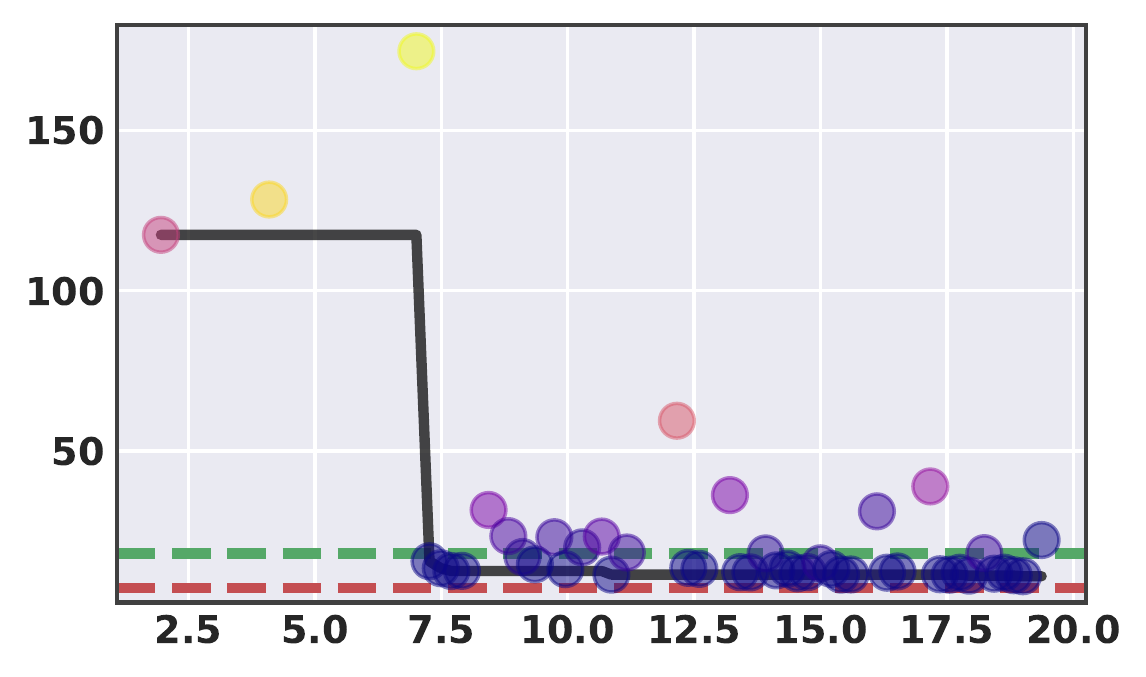}
        & \includegraphics[width=4.7cm]{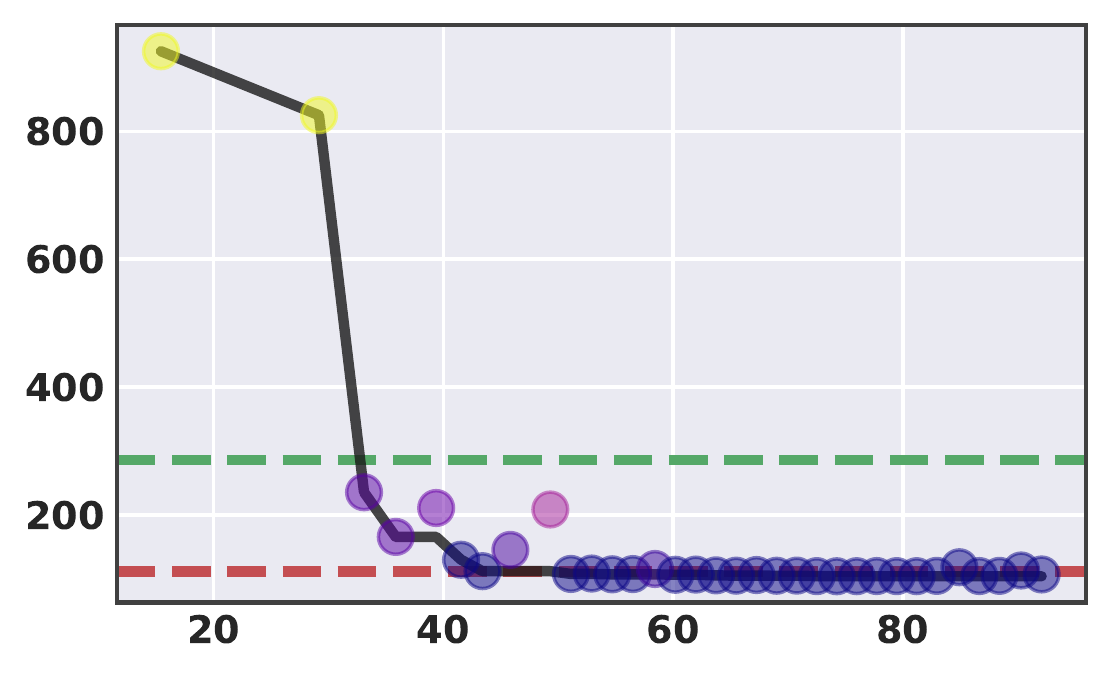} & \includegraphics[width=3cm]{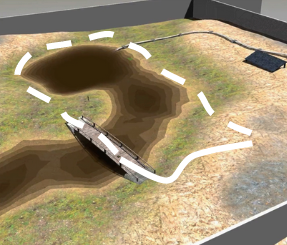} & \rotatebox{90}{\small \env{Sim-F}} \\ %
        \multicolumn{6}{c}{\includegraphics[width=5cm]{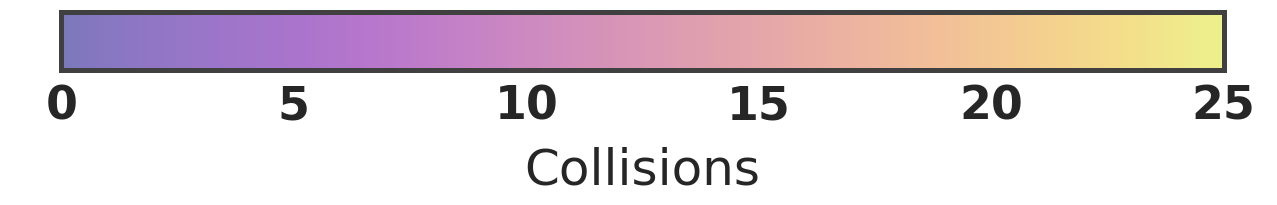}}
        
    \end{tabular}
    \caption{Examples of high-speed driving in diverse, previously unseen environments (outer) by autonomous practicing. \SysName improves lap times (inner; best-so-far shown in black) given a slow \emph{demo lap} (green) and achieves near-expert lap times (red) in under 40 minutes.}
    \label{fig:environments}
\end{figure*}

\section{Faster Lap Times with \SysName}
\label{sec:experiments}

In this section we present an experimental evaluation of \SysName in a variety of real-world and simulated environments. 
We consider several metrics to analyze the peak performance, as well as cumulative metrics during practice. The time-to-first-lap (T2F) represents the time taken to complete the first lap, starting from scratch. We track the best lap time achieved during training as well as the median time of last five laps completed to capture the converged behavior. Additionally, we list the median collisions in the last five laps to capture safety. To contextualize our results, we provide timing for laps driven in each environment by human drivers watching the robot from a third-person view (``Human Expert''), as well as the duration of the ``slow demo'' lap.

Note that we used the \emph{same hyperparameters} (i.e., network architecture, learning rate, reward coefficients) for all experiments discussed here, both in the real-world and simulation. See the appendix for a full list of hyperparameter values. %

\subsection{Real-World Deployment}
\label{sec:results-real}
We deploy \SysName in three \emph{previously unseen} environments to demonstrate autonomous practicing in tightly constrained spaces. Before the start of training, we manually drive the robot around the course for a slow lap to define the rough layout of the track. This lap is used in two ways: (i) to generate a sequence of sparse checkpoints $\{c_i\}_{i=1}^{n_c}$ for the practicing FSM described in Sec.~\ref{sec:methods-fsm}, and (ii) to provide a slow-speed demonstration for off-policy online actor-critic updates as described in Sec.~\ref{sec:methods-online}. The environments described below are shown in Fig.~\ref{fig:environments}, with additional details and visualizations in the appendix.

\env{Indoor-A} represents a large loop (70 meters in length) through the interior of a carpeted building with glass walls and many open corridors. The course is defined by a sequence of $n_c=4$ checkpoints spaced roughly 15-20 meters apart.

\env{Indoor-B} is a significantly larger course ($\sim$120 meters in length) with multiple obstacles, defined by $n_c=4$. The floor of this environment is a tiled and has very low friction, frequently causing dynamical effects such as over/understeer during cornering.

\env{Indoor-C} is a small but challenging indoor race course with two tight ``hairpin'' turns, taken at nearly the maximum steering angle and a tight ``chicane'' (a right-left sequence). Mastering this environment requires the robot to discover fast ``racing lines'' that minimize unnecessary steering, and carrying a high speed through the turns. This course is designated by $n_c=3$ checkpoints. We extensively compare \SysName to alternative baselines in this environment.

\env{Outdoor-D} is a medium-scale (60 meter loop) outdoor course around a building. In addition to straightforward obstacle avoidance with the building, a tree, and a nearby table, there are several patches of tall grass which tend to slow down the robot's motion. A successful policy should avoid the tall grass to the maximum extent possible, staying near paths where the grass is shorter and keeping to the left of the tree when it passes to avoid the grass on the right.

\env{Outdoor-E} is a large-scale (120 meter loop) outdoor course between a dense grove of trees on one side and a tree and several fallen logs on the other side. The ground near the trees is covered in leaves, sticks, and other loose material, causing highly speed-dependent steering characteristics.

\begin{figure*}
    \centering
    \includegraphics[width=\linewidth]{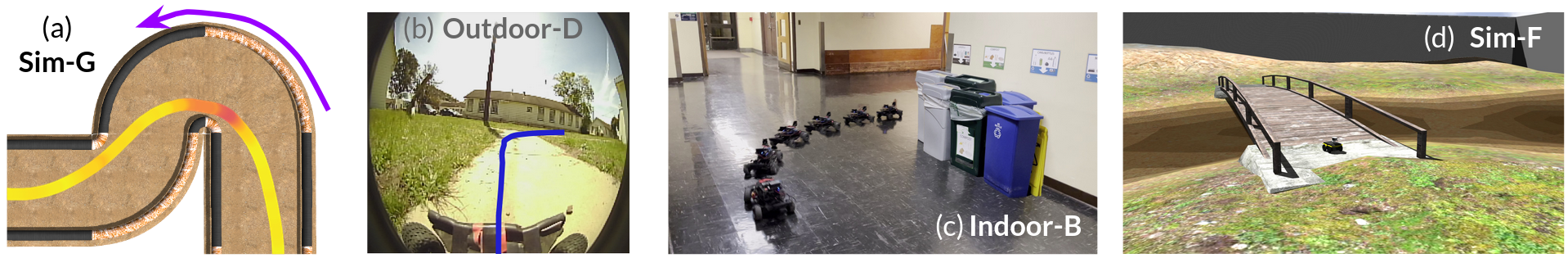}
    \caption{\textbf{Emergent behaviors with \SysName.} We find that directly minimizing lap times with RL results in the emergent discovery of a ``racing line'', braking as late as possible to maintain speed in and out of tight corners (a) and drifting on slick surfaces (c). In a real outdoor environment \env{Outdoor-D}, the robot learns to drive on and near paths to avoid being slowed by tall grass (b). In a challenging simulated outdoor environment \env{Sim-F}, the robot infers that the bridge is faster than driving through mud via visual correlation, learning a fast driving policy (d). \emph{All the above graphics are rendered using telemetry from our real-world (b, c) and simulated (a, d) experiments.} Please see supplemental material for videos of these emergent behaviors.}
    \label{fig:emergent}
\end{figure*}

\begin{table}
\centering
\setlength{\tabcolsep}{4.8pt}
{\footnotesize
\begin{tabular}{lccccccc}
\toprule
& \textbf{T2F} & \multicolumn{4}{c}{\bf Lap Times (s)} & \multicolumn{1}{c}{\bf \# Collisions} \\
\cmidrule(lr){3-6} \cmidrule(lr){7-7}
\textbf{Env.} & \textbf{(s)} & \textbf{Best} & \textbf{Median} &\textbf{\human{Demo}} & \textbf{\human{Expert}$^\dagger$}  & \textbf{Median} \\ \midrule
\env{Indoor-A}  & 114 & 32.7 & 39.0 & \human{54} & \human{25}  & 0 \\ %
\env{Indoor-B}  & 225 & 44.2 & 65.7 & \human{70} & \human{43} & 3  \\ %
\env{Indoor-C}  & 117 & 10.9 & 11.7 & \human{17} & \human{7} & 0 \\ %
\midrule
\env{Outdoor-D}  & 196 & 17.1 & 22.7 & \human{43} & \human{18}  & 0 \\ %
\env{Outdoor-E}  & 282 & 62.1 & 94.0 & \human{160} & \human{40} & 3  \\ %
\midrule
\env{Sim-F} & 925 & 104.1 & 107.0 & \human{286} & \human{112} & 0 \\ %
\env{Sim-G} & 174 & 18.0 & 18.1 & \human{36} & \human{19} & 0\\ %
\bottomrule
\end{tabular}}
\caption{\textbf{Summary of experiments:} \SysName can consistently learn aggressive driving policies in environments of varying difficulty levels, improving over the \emph{demo lap} by over 40\% and achieving lap times within 5\% of the expert, using only egocentric observations. \emph{$^\dagger$Note that the expert has access to privileged third-person observations.}}
\label{tab:real-summary}
\end{table}

Table \ref{tab:real-summary} and Fig.~\ref{fig:environments} summarize the performance of our system in these environments. \SysName is able to consistently improve over the low-speed demonstration lap in a handful of laps, and nearly match human performance in \env{Indoor-B} and \env{Outdoor-D} in 30 minutes of real-world practice, without any human interventions. As training progresses, the achieved lap times continue to decrease, with the path taken by the robot becoming more optimized as a secondary effect of optimizing speed Fig.~\ref{fig:real-evolving}.

\MyPara{Emergent behaviors}: We find that directly maximizing the reward for reaching the next checkpoint as quickly as possible (Eqn.~\ref{eq:reward}) leads to emergent behaviors in our system, visualized in Fig.~\ref{fig:emergent}. The system learns the concept of a ``racing line'', finding a smooth path through the lap and maximizing its speed through tight corners and chicanes (a). This can be seen in the speed profile through a tight corner in Fig.~\ref{fig:emergent}(a), where the robot learns to carry its speed into the \emph{apex}, then brakes sharply to turn and accelerates out of the corner, to minimize the driving duration. In Fig.~\ref{fig:emergent}(c) with a low-friction surface, the policy learns to over-steer slightly when turning, drifting into the corner to achieve fast rotation without braking during the turn (c). In outdoor environments, the learned policy is also able to distinguish ground characteristics, preferring smooth, high-traction areas on and around concrete paths in \env{Outdoor-D} over areas with tall grass that impedes the robot's motion. Please see our website for videos of our system practicing and driving aggressively. 

\subsection{Simulated Environments}
\label{sec:results-sim}

We also test our approach in two simulated environments --- a rally racing track, and a large, visually challenging off-road environment ---
to evaluate \SysName in more diverse settings. For these experiments we use a simulated Clearpath Jackal robot with identical observation space to our robot. Unlike our RC car-based platform, the Jackal uses a differential drive rather than an Ackermann steering setup; due to this difference the action space consists of linear and angular velocity targets rather than linear velocity and steering.

\env{Sim-F} is a large, complex world derived from Clearpath's simulation environments~\cite{clearpath}. In this environment the robot must navigate around a large pool of mud, which shows non-binary traversability --- greatly limiting the robot’s dynamics and maximum speed --- and should be avoided when possible. A bridge allows the robot to bypass the mud, but is narrow and can be identified visually. Our method successfully learns a high-performance policy in this environment, successfully correlating the mud's appearance with a low rewards, and selecting the optimal path after the first few laps of trial and error (see Fig.~\ref{fig:emergent}), achieving super-human lap times in under 10 laps (see Tab.~\ref{tab:real-summary}). This environment is particularly challenging since the mud can cause the robot to get irrecoverably stuck in a slow-speed zone; all alternative baselines and ablations failed to solve this task.

\env{Sim-G} is a smaller-scale dirt track with sharp turns and chicanes, much like \env{Indoor-C}, making it a particularly interesting environment to study the emergence of \emph{racing lines} and agile maneuvers. All baselines were able to solve this task to some extent, allowing direct comparison between several alternative approaches to the problem.

\subsection{Comparative Analysis}
\label{sec:results-analysis}

We compare the performance of \SysName against the baselines and ablations mentioned in two environments: \env{Indoor-C} and \env{Sim-G}. we compare our approach to several baselines and ablations, and demonstrate the importance of each of the components of our method: pre-trained visual representations, online RL starting from a slow demo lap, and autonomous recovery behaviors to handle the reset-free environment. Specifically, we consider the following variations and ablations:

\MyPara{Offline RL:} Ablating the online learning aspect, this baseline uses a policy trained purely offline, with access to 15 minutes of \emph{expert} data. Note that this is \emph{more} on-task offline data than is available to \SysName.

\MyPara{No Demo Lap:} Ablating the online learning aspect, this baseline deploys \SysName \emph{without} sampling any data from the low-speed demo lap $\gB_\text{slow}$ (Alg.~\ref{alg:main}, L\ref{algline:slow-sample}).

\MyPara{No Pre-Training:} Ablating offline pre-training, this baseline uses networks initialized from scratch and trained online using DrQ~\cite{kostrikov2021image}, a state-of-the-art pixel-based RL algorithm.

\MyPara{ImageNet Pre-Training:} Ablating task-specific pre-training, this baseline uses the same encoder structure, but trained instead for image classification on ImageNet~\cite{deng2009imagenet, parisi2022unsurprising} for extracting visual features.

\MyPara{No Pseudo-Resets:} Ablating the FSM, this baseline deploys \SysName \emph{without} the benefit of scripted pseudo-resets when it is stuck, requiring the robot to learn recovery behavior.

\MyPara{State-Based:} This variation replaces visual observations with \emph{privileged} state estimates in the form of an approximate 2D pose (measured by onboard visual-inertial odometry).

\begin{table}
\centering
{\footnotesize
\begin{tabular}{lccccc}
\toprule
& & \multicolumn{2}{c}{\bf Lap Times (s)} & \textbf{\# Collisions} \\
\cmidrule(lr){3-4} \cmidrule(lr){5-5}
 & \textbf{T2F (s)} & \textbf{Best} & \textbf{Median} & \textbf{Median} \\ \midrule
\multicolumn{5}{l}{\textcolor{darkgray}{\textbf{Real-World}} (\env{Indoor-C})} \\
State-Based & 274 & 12.7 & 18.8 & 3 \\
Offline RL~\cite{shah2022offline} & $\infty$ & -- & -- & -- \\ %
No Pre-Training & 233 & 12.7 & 20.0 & 1  \\
ImageNet Pre-Training & \textbf{49.8} & 19.7 & 29.7 & 1  \\
No Demo Lap & 239 & 16.0 & 62.6 & 12 \\
\ours \SysName (Ours) & \textbf{117} & \textbf{10.9} & \textbf{13.3} & \textbf{0} \\
\midrule
\human{Human FPV} & \human{--} & \human{11.1} & \human{14.4} & \human{2}  \\
\human{Human Oracle$^\dagger$} & \human{--} & \human{7.3}  & \human{8.8} & \human{0} \\
\midrule \midrule
\multicolumn{5}{l}{\textcolor{darkgray}{\textbf{Simulation}} (\env{Sim-G})} \\
State-Based & 222 & 18.9 & 26.2 & 0\\
No Demo Lap & 665 & 19.6 & 22.2  & 0\\
No Pre-Training & 375 & \textbf{17.8} & 18.4 & 1\\
No Pseudo-Resets & 405 & 21.7 & 25.1  & 0\\
\ours \SysName (Ours)  & \textbf{173} & \textbf{18.0} & \textbf{18.1} & 0 \\
\midrule
\human{Human Oracle$^\dagger$} & \human{--} & \human{18.6} & \human{18.9}  & \human{0}\\ %
\bottomrule
\end{tabular}}
\caption{\textbf{Comparing to baselines} in real-world and simulated environments, \SysName achieves consistently lower time-to-first lap (T2F), best and median lap times, and minimum collisions. Notably, \SysName can outperform a system with access to privileged state estimates, and a human driving from first-person view (FPV). In both environments, \SysName achieves close-to-oracle$^\dagger$ driving performance, i.e., an expert human driving with \emph{privileged} third-person observations of the robot, shown in \human{gray}. Offline RL failed to complete a single lap in \env{Indoor-C}, likely due to its inability to adapt the learned policies on-the-fly in novel environments.}
\label{tab:comparisons-all}
\end{table}

\begin{figure}[ht]
    \centering
    \includegraphics[width=\columnwidth]{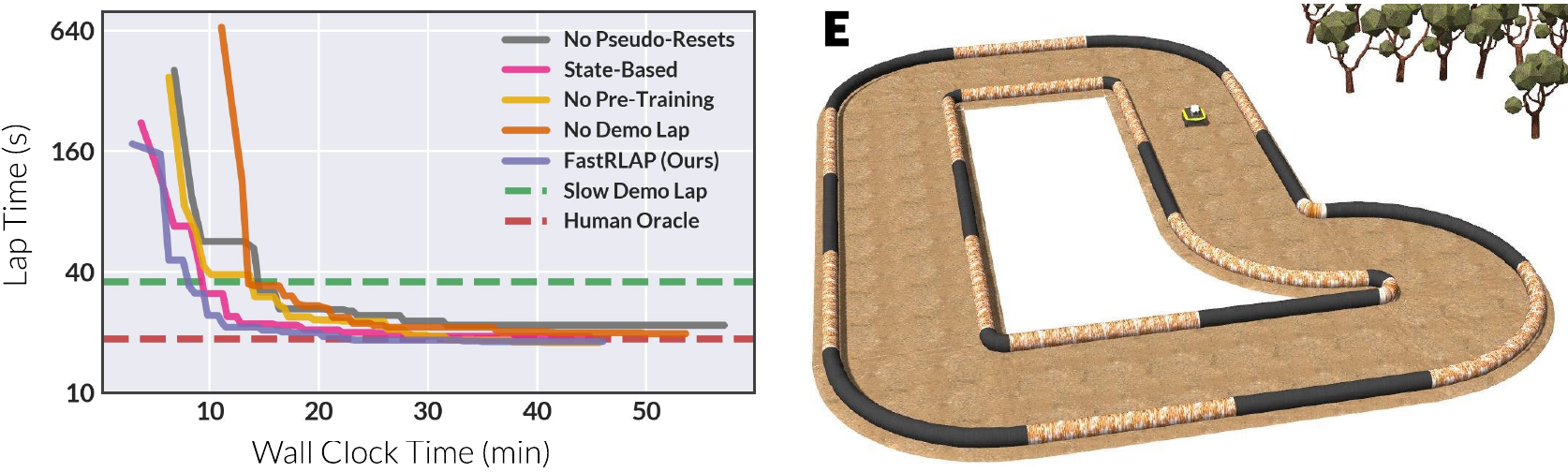}
    \caption{Progression of running minimum lap times across different baselines in Tab.~\ref{tab:comparisons-all} in \env{Sim-G}. \SysName achieves expert-level performance in 20 minutes, learning efficiently from pixels as quickly as an agent with privileged state information, while achieving better overall performance by the end of training.}
    \label{fig:sim-baseline-curves}
\end{figure}

In \env{Indoor-C}, our experiments (Tab.~\ref{tab:comparisons-all}) show that \SysName far outperforms the ablation without a demo lap in both time-to-first lap (T2F) and best lap time, while also encountering fewer collisions. Early in the training process, the demonstration helps the system make progress around the course, enabling broad state coverage, which yields more useful exploration early on and eventually leads to better performance.
The offline RL baseline completely failed to produce a usable policy, as it is unable to adapt online to new rich observations.

 Several additional baselines were considered in the simulated \env{Sim-G} environment, summarized in Tab.~\ref{tab:comparisons-all}. The same trends hold, showing that the demo lap is very important to fast learning and achieving a low T2F. Similarly, removing pseudo-resets causes the system to get stuck for extended durations, resulting in a very slow first lap (even with a demonstration!). Fig.~\ref{fig:sim-baseline-curves} shows the progression of the running minimum lap times for each method.

Analyzing the role of pre-training with offline RL, we find that \SysName initialized with a generic ImageNet encoder completes its first lap relatively quickly, achieving a T2F comparable to \SysName. However, its asymptotic performance is comparably poor: its best lap time across training in \env{Indoor-C} is only slightly better than the slow-lap demonstration. This suggests that while extracting general-purpose visual features (e.g., edges and gradients) may be sufficient for low-speed navigation, high-speed navigation requires consideration of task-specific features, such as depth or obstacle detection, that are better learned with task-specific pre-training.

Most surprisingly, learning directly from visual observations outperformed the variation with access to privileged state information in both simulated and real environments. This suggests that the features learned by the pre-trained encoder are \emph{more informative} than simple localization estimates, because they generalize better: an obstacle has similar representation across different positions and environments, while the state-based agent must learn a free-space representation of the entire environment inside its critic function via trial and error, bumping into each object before it can record its existence.

\subsection{Qualitative Analysis}
To better understand if the learned policy is genuinely learning from the visual cues in the environment or simply memorizing a sequence of actions, we consider the value of $Q(s, a)$ for a set of sample images by artificially injecting a range of steering inputs (actions). This enables us to see how the critic network assesses the desirability of different actions based on the visual information in the environment.

We find that the visual observations qualitatively explain the critic outputs, with obstacles corresponding to low assigned values to actions steering towards them and relatively higher values assigned to keeping to less-restrictive areas like paths rather than tall grass. This suggests that there appears to be a correlation between the visual cues in the environment and the actions chosen by the learned policy, suggesting that it is not merely memorizing a sequence of actions but rather learning to make decisions based on the environment's visual features.

\begin{figure}
  \centering
  \begin{minipage}[t]{0.47\columnwidth}
    \includegraphics[width=0.99\columnwidth, trim={0px 0px 30px 0px}, clip]{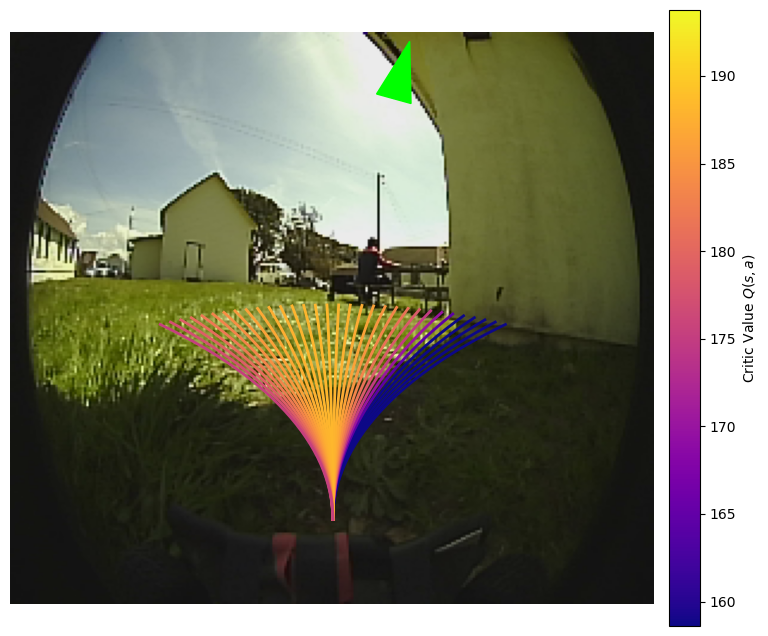}
    \subcaption{The critic assigns very low value to actions that would cause contact with an obstacle.}
  \end{minipage}
  \hfill
  \begin{minipage}[t]{0.47\columnwidth}
    \includegraphics[width=0.99\columnwidth, trim={0px 0px 45px 10px}, clip]{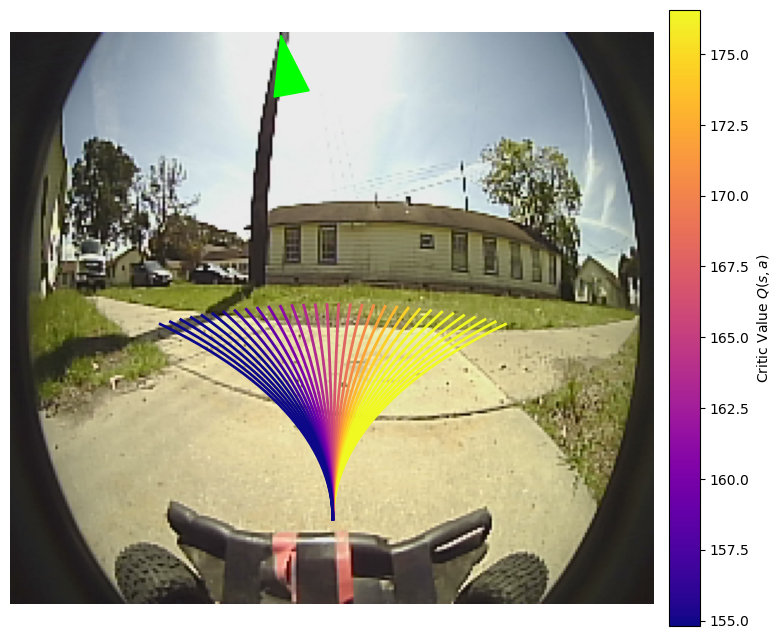}
    \subcaption{The critic suggests turning towards the next checkpoint before the current checkpoint is reached.}
  \end{minipage}
  
  \begin{minipage}[t]{0.47\columnwidth}
    \includegraphics[width=0.99\columnwidth, trim={0px 0px 30px 0px}, clip]{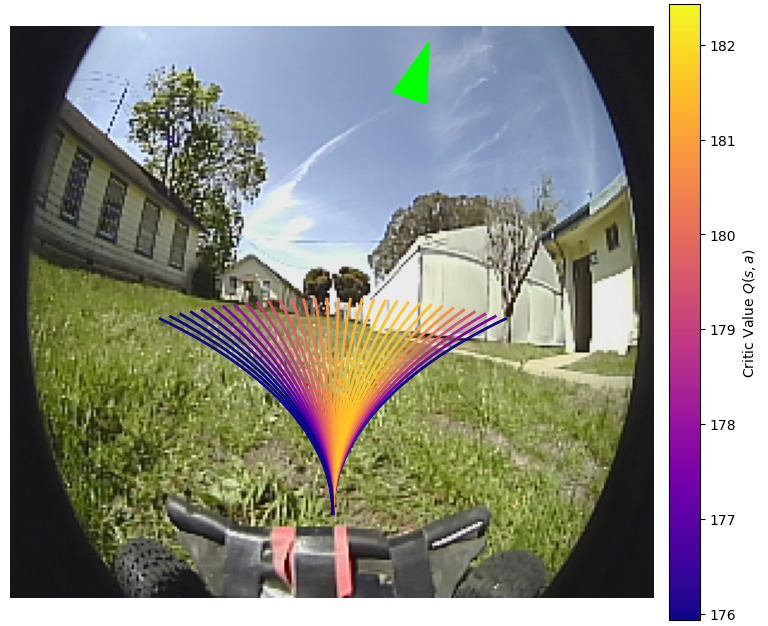}
    \subcaption{In wide-open areas the critic reflects a beeline policy directly towards the next goal.}
  \end{minipage}
  \hfill
  \begin{minipage}[t]{0.47\columnwidth}
    \includegraphics[width=0.99\columnwidth, trim={0px 0px 10px 0px}, clip]{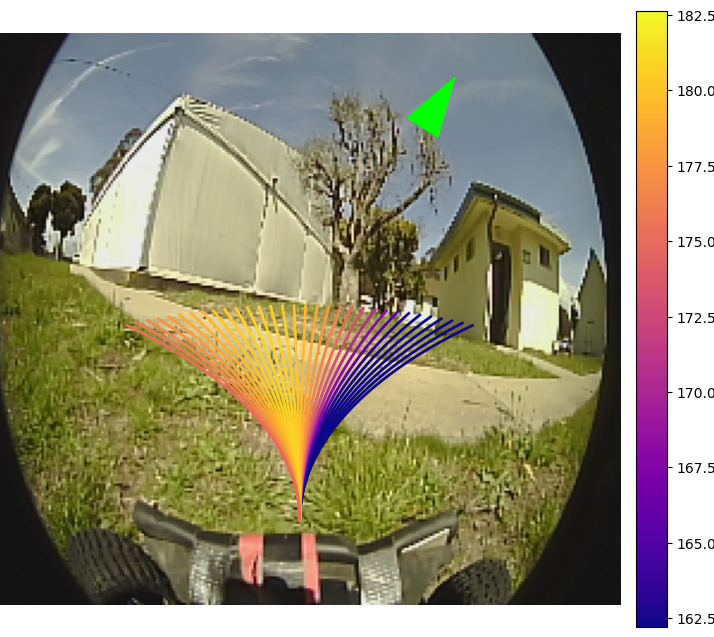}
    \subcaption{The critic prefers turning left rather than the geometric shortest-path on the right to avoid tall grass.}
  \end{minipage}
  \caption{\textbf{Qualitative critic evaluations} on sample images from environment \env{Outdoor-D} against diverse steering actions, represented as (approximate) geometric paths overlaid on the image. Hot (cold) colors represent high (low)-value actions. The green arrow points in the direction of the next checkpoint.}
  \label{fig:critic_analysis}
\end{figure}

\newpage
\section{Discussion} 
\label{sec:discussion}

We presented a system for learning high-speed driving with reinforcement learning from rich observations, practicing autonomously in the real world. Our approach uses representations from prior data to initialize the policy, followed by sample-efficient online RL paired with a checkpoint-based navigation strategy to recover autonomously from collisions and continue practicing. Although deep RL is often believed to be inefficient and difficult to use in the real world, we demonstrate that with appropriate pre-training and several important design decisions, our system can actually learn effective driving strategies in under 20 minutes of real-world training. This result may seem quite surprising when viewed in contrast to prior work that uses simulated data~\citep{loquercio2020deep}, or hundreds of hours of training~\citep{wijmans2020ddppo}, and we believe it provides strong validation that deep RL can indeed be a viable tool for learning real-world policies even from raw images, when combined with appropriate pre-training and implemented in the context of an autonomous training framework.

A qualitative investigation of the policies learned by our system also reveals interesting emergent behavior. Although we bootstrap training with prior data (in other domains and from other robots) and a single slow demonstration lap, the learned policies exhibit skills like drifting, selecting for high-speed terrain, and maintaining a racing line which deviate significantly from the behaviors seen in the prior data. Thus, the online RL process not only serves to robustify previously seen behavior, as observed in prior work incorporating offline data into real-world RL~\citep{kumar2022pre}, but also acquires new emergent behaviors building on the foundation established by the prior data. At the same time, our ablations establish the importance of task-relevant pre-training, supporting the notion that representations learned from diverse robot navigation data serve as an effective foundation for downstream skill learning, resembling the importance of self-supervised pre-training objectives in vision and NLP~\citep{devlin2018bert,chen2020improved}.

While our system enables highly effective image-based driving, it does have a number of limitations. First, the current implementation still relies on a state estimator to provide a vector to the next checkpoint. This limitation could be addressed in future work by using image goals rather than position goals, as has been demonstrated in prior work on long-horizon navigation~\citep{shah_viking_2022}. Second, although our system provides for autonomous recovery from collisions, it does not actually account for \emph{safety} explicitly: that is, the system aims to avoid collisions because it leads to task failure, but does not perform any special safety-aware handling of unfamiliar inputs to, for example, slow down or proceed cautiously in unfamiliar situations. This kind of safety handling would likely be essential in more complex open-world settings or with larger vehicles that can seriously damage themselves in a collision. We hope that addressing these limitations will enable RL-based systems to learn complex and highly performant navigation skills in a wide range of domains, and we believe that our work can provide a stepping stone toward this.

\section*{Acknowledgments}
This research was partially supported by DARPA RACER, ARL DCIST CRA W911NF-17-2-0181, the National Science Foundation through IIS-2150826, and the Office of Naval Research. The authors would like to thank Alejandro Escontrela, Noriaki Hirose, and Philippe Hansen-Estruch, for their help with running experiments and providing baseline implementations.

\bibliographystyle{IEEEtran}
\bibliography{references}

\appendix
\FloatBarrier

\subsection{Hyperparameters}
\label{sec:app:hyperparams}

\begin{figure}
    \centering
    \includegraphics[width=0.8\columnwidth]{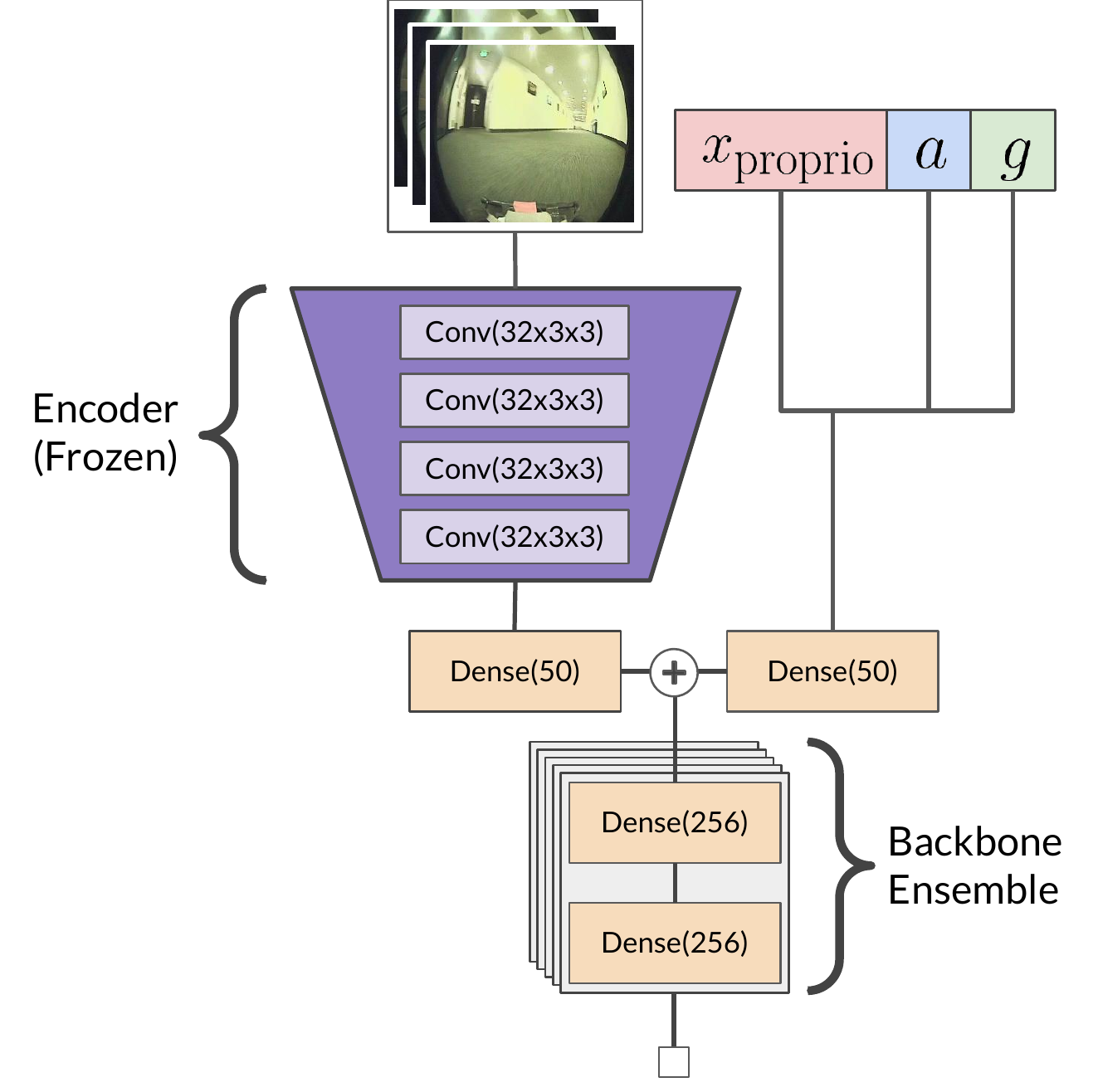}
    \caption{Network architecture for the critic. Actor (and, for IQL, value) architectures are identical, but with only one backbone rather than an ensemble. Prorpioceptive information is concatenated with the action and the goal and fed through a dense layer, concatenated with the output of a convolutional encoder applied to a sequence of three camera images, and fed through a 2-layer MLP.}
    \label{fig:network_architecture}
\end{figure}

All our experiments use the \emph{same} set of hyperparameters and there is no environment-specific tuning used in the results presented in the paper.
See Tab.~\ref{tab:hparams} for a list of hyperparameters and Fig.~\ref{fig:network_architecture} for a detailed network architecture.

\begin{table}[H]
\centering
\small
\begin{tabular}{lll}
\toprule
\textbf{Category} & \textbf{Hyperparameter} & \textbf{Value} \\ \midrule
Actor/Critic & Actor learning rate & 3e-4 \\
& Critic learning rate & 3e-4 \\
& Temperature learning rate & 3e-4 \\
& Actor network architecture & 2x256 \\
& Critic network architecture & 2x256 \\
& Initial target entropy & -3 \\
& Entropy decay rate & 1e-5 \\
& Critic ensemble size & 10 \\ \midrule
MDP/System & Discount factor & 0.99 \\
& Time step & 0.1s \\ 
& Velocity target range (m/s) & $[0.5, 3.5]$ \\
& Servo target range (rad) & $[-0.5, 0.5]$ \\
& $C_{\textrm{collide}}$ (real only) & $0.2s^2/m$ \\
& $C_{\textrm{stuck}}$ & -10 \\
& Squashing range $\delta$ & 0.2 \\ \midrule
Encoder & Layer count & 4 \\
& Convolution size & 3x3 \\
& Stride & 2 \\
& Hidden channels & 32 \\ \midrule
IQL & Expectile & 0.7 \\
& Value network structure & same as critic \\ \bottomrule
\end{tabular}
\caption{List of hyperparameters used throughout experiments}
\label{tab:hparams}
\end{table}

\begin{figure*}
\footnotesize
    \centering
    \begin{tabular}{|l|l|rrrr|rrrrrr|}
    \hline
     & & \multicolumn{4}{c|}{Lap time (s)} & \multicolumn{6}{c|}{\# collisions}\\
    Course name & Method & \multicolumn{1}{c}{first} & \multicolumn{1}{c}{best} & \multicolumn{1}{c}{median} & \multicolumn{1}{c|}{median (5)} & \multicolumn{1}{c}{best} & \multicolumn{1}{c}{total} & \multicolumn{1}{c}{mean} & \multicolumn{1}{c}{median} & \multicolumn{1}{c}{mean (5)} & \multicolumn{1}{c|}{median (5)}\\
    \hline
    \hline
    \texttt{Real-A} & FastRLAP & 114.31 & 32.74 & 49.51 & 38.99 & 0 & 82 & 2.48 & 2 & 2.00 & 0\\
    \hline
    \texttt{Real-B} & FastRLAP & 224.70 & 44.21 & 73.71 & 65.72 & 0 & 135 & 7.50 & 7 & 4.20 & 3\\
    \hline
    \texttt{Real-C} & States & 274.42 & 12.70 & 50.13 & 18.88 & 0 & 190 & 12.67 & 4 & 3.40 & 2\\
    \texttt{Real-C} & Ours & 117.38 & 10.90 & 13.27 & 11.69 & 0 & 126 & 2.74 & 0 & 0.00 & 0\\
    \texttt{Real-C} & No slow lap & 239.21 & 16.01 & 64.51 & 62.62 & 0 & 206 & 22.89 & 12 & 8.80 & 11\\
    \texttt{Real-C} & Human Oracle & 54.40 & 7.21 & 10.08 & 8.79 & 0 & 7 & 1.00 & 0 & 0.00 & 0\\
    \texttt{Real-C} & ImageNet encoder & 49.75 & 19.66 & 30.20 & 21.12 & 0 & 168 & 6.46 & 1 & 0.00 & 0\\
    \texttt{Real-C} & No pretraining & 232.79 & 12.70 & 21.60 & 19.99 & 0 & 174 & 9.67 & 1 & 1.80 & 1\\
    \hline
    \texttt{Sim-D} & FastRLAP & 925.08 & 104.19 & 107.00 & 107.00 & 0 & 157 & 4.76 & 0 & 0.00 & 0\\
    \hline
    \texttt{Sim-E} & Blind & 222.17 & 18.89 & 21.70 & 19.50 & 0 & 127 & 1.18 & 0 & 1.20 & 0\\
    \texttt{Sim-E} & States & 222.17 & 19.10 & 23.69 & 26.20 & 0 & 113 & 1.30 & 0 & 1.00 & 1\\
    \texttt{Sim-E} & FastRLAP & 174.30 & 17.99 & 19.10 & 18.10 & 0 & 48 & 0.42 & 0 & 0.00 & 0\\
    \texttt{Sim-E} & No slow lap & 665.04 & 19.70 & 23.33 & 22.20 & 0 & 90 & 0.95 & 0 & 0.00 & 0\\
    \texttt{Sim-E} & No pretraining & 375.26 & 17.80 & 21.90 & 18.39 & 0 & 100 & 1.14 & 0 & 0.00 & 0\\
    \texttt{Sim-E} & No pseudo-resets & 405.02 & 21.70 & 26.31 & 25.10 & 0 & 156 & 1.64 & 0 & 0.20 & 0\\
    \hline
    \end{tabular}
    \captionof{table}{Detailed statistics of all runs}
    \label{tab:detailed-stats}
\end{figure*}

\begin{figure*}
    \centering
    \includegraphics[width=\linewidth]{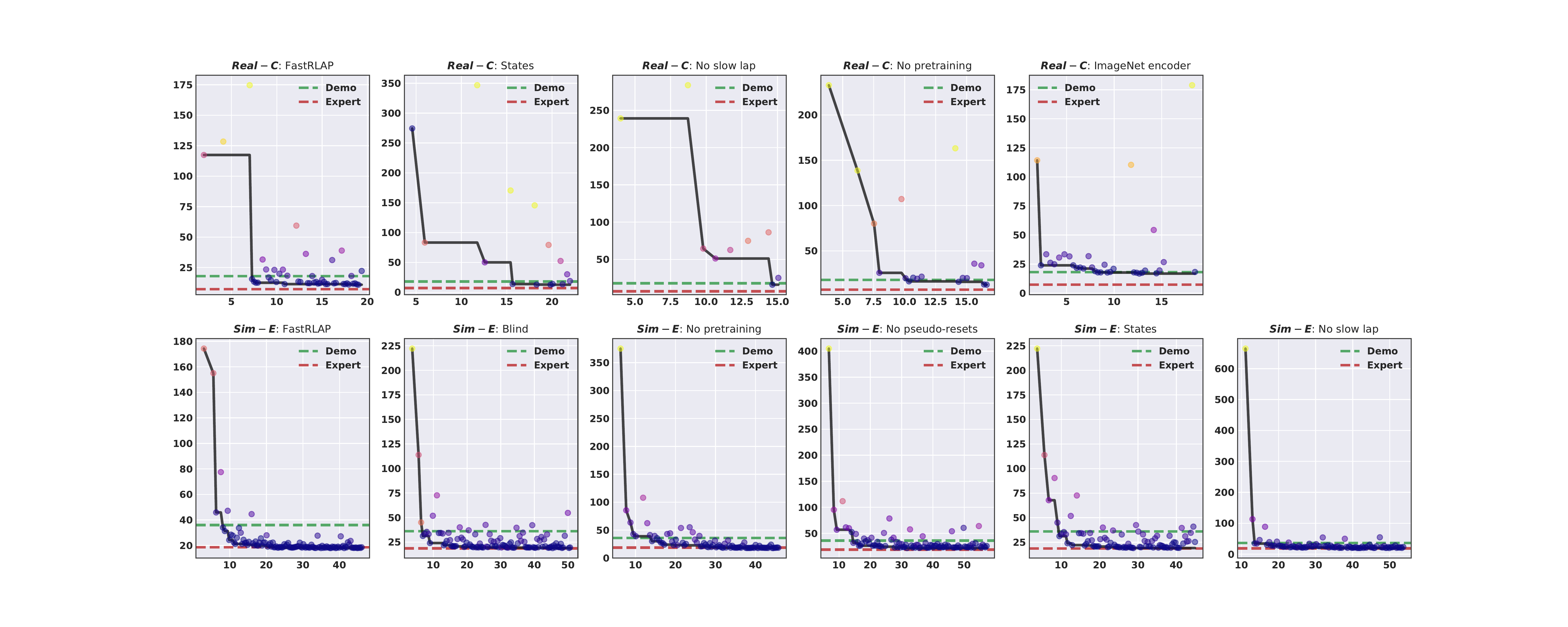}
    \captionof{figure}{Detailed laptime progression charts for all baselines}
    \label{fig:laptime-details}
\end{figure*}

\subsection{Additional Experimental Details}
In addition to the metrics presented in the main paper, please see Tab.~\ref{tab:detailed-stats} for a list of additional metrics per experiment, and Fig.~\ref{fig:laptime-details} for lap time progression charts for each experiment. FastRLAP outperforms baselines in \emph{all} environments, as measured by any suitable metric --- time-to-first lap, best lap time, mean/median lap times, and minimum number of average and best-case collisions.

\subsection{Implementation Details}
The overall system was implemented using ROS 1 Noetic Ninjemys, with the inference and training code using JAX. We transfered tensors between the components of our system (new data from the robot to the workstation, and parameteres from the workstation to the robot) using ROS messages.

\subsection{System Details}
\begin{figure*}
    \centering
    \includegraphics[width=0.6\linewidth]{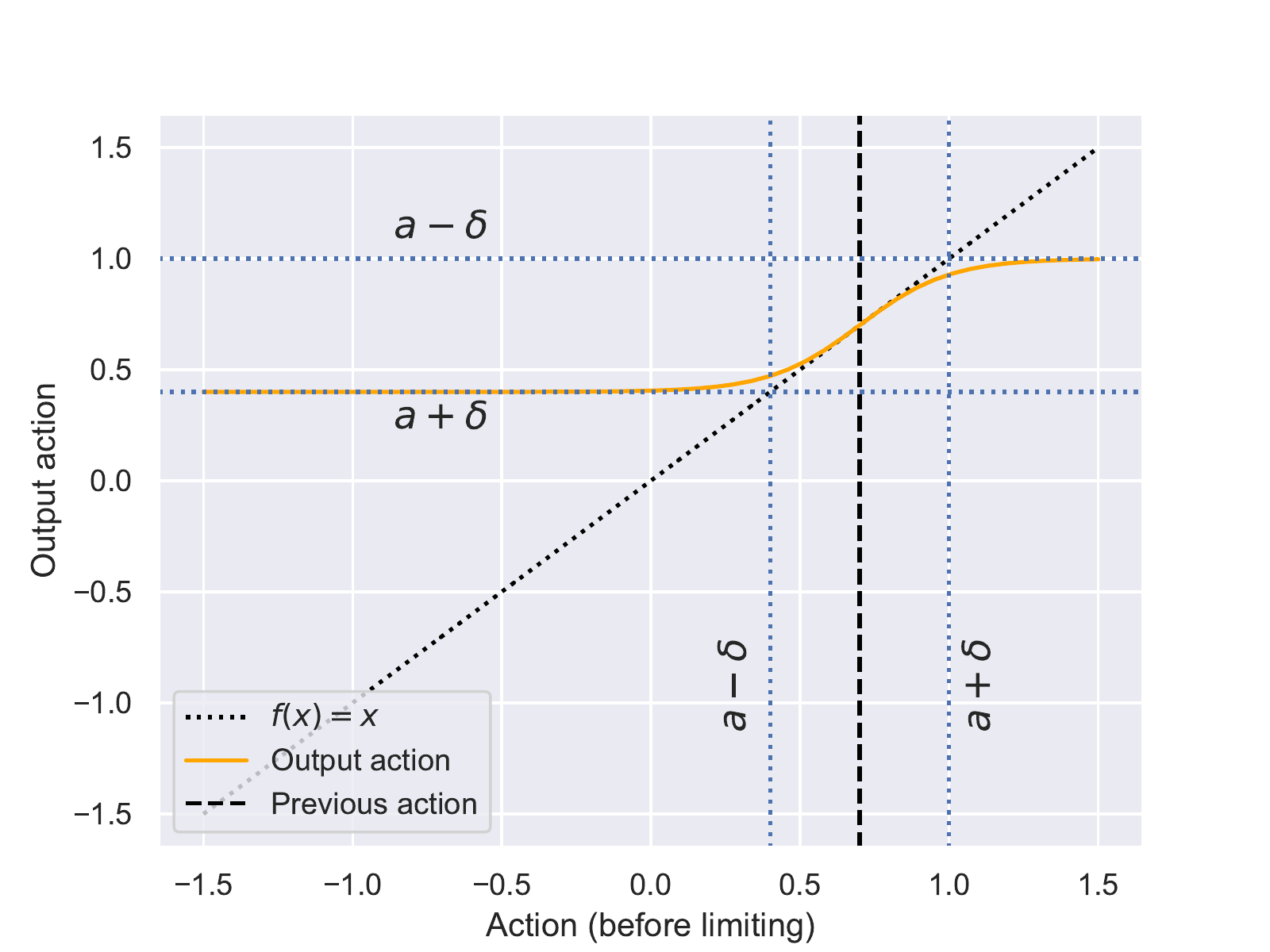}
    \caption{Our activation function (actor output only) ensures action continuity.}
    \label{fig:tanh_squashing}
\end{figure*}

As mentioned in the paper, we squash the output of the actor dynamically to a range around the previous action $a_{\textrm{prev}}$ to enforce continuity in output actions by constraining them to be no more than some positive constant $\delta$ away from the previous action in each dimension. We choose our activation function so that the interval $[a_{\textrm{prev}} - \delta, a_{\textrm{prev}} + \delta]$ is roughly mapped to itself. More precisely, we ensure that our activation $f_{a_{\textrm{prev}}, \delta}(x)$ is accurate to first-order Taylor expansion around the previous action. This yields the following activation function applied to the output of the actor:
\[f_{a_{\textrm{prev}}, \delta}(x) = \tanh\left(\frac{(x - a_{\textrm{prev}})}{\delta}\right)\delta + a_{\textrm{prev}}\]
Figure \ref{fig:tanh_squashing} shows a graphical depiction of this activation function.

\subsection{Simulation Environment}
The Clearpath Jackal used for simulations differs from the real environments primarily in its action space, which (as a differential drive robot) allows turning in place. We limit the linear velocity actions of the robot to $[-1, 2]$ and the angular velocity actions to $[-1.0, 1.0]$. Simulated position measurement is provided in lieu of the RealSense tracker for determining relative goal locations.

\subsection{Code Release}
Please see \href{https://sites.google.com/view/fastrlap}{\bf sites.google.com/view/fastrlap} for the training and robot-side inference code as well as modified simulation environments.

\end{document}